# Towards deep observation: A systematic survey on artificial intelligence techniques to monitor fetus via Ultrasound Images

MAHMOOD ALZUBAIDI[1], MARCO AGUS[1], KHALID ALYAFEI[2 3], KHALED A ALTHELAYA[1], UZAIR SHAH[1], ALAA A. ABD-ALRAZAQ[1], MOHAMMED ANBAR[3], MICHEL MAKHLOUF [2], AND MOWAFA HOUSEH[1]

[1] College of Science and Engineering, Hamad Bin Khalifa University, Doha, Qatar; [2] Weil Cornell Medical College-Qatar; [3]Sidra Medical and Research Center, Doha, Qatar; [3] National Advanced IPv6 Centre, University Sains Malaysia, Penang, Malaysia

## Abstract

**Objective:** Developing innovative informatics approaches aimed to enhance fetal monitoring is a burgeoning field of study in reproductive medicine. Over the last decade, several reviews have been conducted regarding Artificial intelligence (AI) techniques to improve pregnancy outcomes. They are limited by focusing on specific data (e.g., electronic health records) or a mother's care during pregnancy. This survey aims to explore how artificial intelligence (AI) can assist with fetal growth monitoring via Ultrasound (US) image.

**Materials and methods:** We performed a systematic survey guided by the Joanna Briggs Institute Reviewer Manual. We reported our findings using the guidelines for PRISMA (Preferred Reporting Items for Systematic Reviews and Meta-Analysis). We conducted a comprehensive search of eight medical and computer science bibliographic databases, including PubMed, Embase, PsycINFO, ScienceDirect, IEEE explore, ACM Library, Google Scholar, and the Web of Science. We retrieved studies published between 2010 to 2021. The search terms were chosen based on the target intervention (i.e., AI and ultrasound images) and target population (i.e., fetal/fetus). Publications written in English that converged on AI techniques used to monitor fetal care during pregnancy regardless of target fetal disease or anatomy were included. Data extracted from studies were synthesized using a narrative approach.

**Results:** Out of 1269 retrieved studies, we included 107 distinct studies from queries that were relevant to the topic in the survey. We found that 2D ultrasound images were more popular (n = 88) than 3D and 4D ultrasound images (n = 19). Classification is the most commonly used method (n = 42), followed by segmentation (n = 31), classification integrated with segmentation (n = 16) and other miscellaneous such as object-detection, regression and reinforcement learning (n = 18). The most common areas that gained traction within the pregnancy domain were the fetus head (n = 43), then fetus body (n=31), fetus heart (n=13), fetus abdomen (n=10), and lastly the fetus face (n=10). In the most recent studies, deep learning techniques were primarily used (n = 81), followed by machine learning (n=16), artificial neural network (n=7), and reinforcement learning (n=2). We did not find studies that developed AI-based applications at any medical center.

**Conclusions:** Overall, we found that AI techniques played a crucial role in predicting particular fetal diseases and identifying fetus anatomy structures during pregnancy. More research is required to validate this technology from a physician's perspective, such as pilot studies and randomized controlled trials on AI and its applications in a hospital setting. Additional research is required to focus in how radiologists can utilize AI technique because when AI is integrated into the clinical process as a tool to assist clinicians, more accurate and reproducible radiological assessments can be made.

## 1. Introduction

### 1.1 Background

Artificial Intelligence (AI) is a broad discipline that aims to replicate the inherent intelligence shown by people via artificial methods [1]. Recently, AI techniques have been widely utilized in the medical sector [2]. Historically, AI techniques were stand-alone systems with no direct link to medical imaging. With the development of new technology, the idea of 'joint decision-making' between people and AI offers the potential of boosting high performance in the area of medical imaging [3].

*Correspondence to*: Dr. Mowafa Househ, College of Science and Engineering, Hamad Bin Khalifa University, Doha, Qatar
Tel.: +xxxxxxxxxxxxxxxxxxxxx; E-mail: mhouseh@hbku.edu.qa

In computer science, Machine Learning (ML), Deep Learning (DL), Artificial Neural Network (ANN) and Reinforcement Learning (RL) are subset techniques of AI that are used to perform different tasks on medical images such as classification, segmentation, object identification, and regression [4]–[6]. Diagnosis using computer-aided detection (CAD) has moved towards becoming AI automated process in the medical images[7], which include most of the medical imaging data such as (X-ray radiography, Fluoroscopy, MRI, Medical ultrasonography or ultrasound, Endoscopy, Elastography, Tactile imaging, and Thermography) [8]. However, digitized medical images come with a plethora of new information, possibilities, and challenges. Therefore, AI techniques are able to address some of these challenges by showing impressive accuracy and sensitivity in identifying imaging abnormalities. These techniques promise to enhance tissue-based detection and characterization, with the potential to improve diagnoses of diseases [9].

At present, the use of AI techniques in medical images has been discussed in depth across many medical disciplines, including identifying cardiovascular abnormalities, detecting fractures and other musculoskeletal injuries, aiding in diagnosing neurological diseases, reducing thoracic complications and conditions, screening for common cancers, and many other prognoses and diagnosis tasks [7], [10]–[13]. Furthermore, AI techniques have shown the ability to provide promising findings when utilizing prenatal medical images, such as monitoring fetal development at each stage of pregnancy, predicting a pregnant placenta's health, and identifying potential complications [14]–[17]. AI techniques may assist with detecting several fetal diseases and adverse pregnancy outcomes with complex etiology and pathogenesis such as amniotic band syndrome, congenital diaphragmatic hernia, congenital high airway obstruction syndrome, fetal bowel obstruction, gastroschisis, omphalocele, pulmonary sequestration, and sacrococcygeal teratoma [18]–[20]. Therefore, more research is needed to understand the role of utilizing AI techniques in the early stage of pregnancy both to prevent and reduce unfavorable outcomes as well as provide an understanding of fetal anomalies and illnesses throughout pregnancy and for drastically reducing the need of more invasive diagnostic procedures that may be harmful for the fetus [21].

Various imaging modalities (e.g., Ultrasound, MRI, and computed tomography (CT)) are available and can be utilized by AI technique during pregnancy. [6] In medical literature, ultrasound imaging has become popular and is used during all pregnancy trimesters. Ultrasound is crucial for making diagnoses, along with tracking fetal growth and development. Additionally, Ultrasound can provide both precise fetal anatomical information as well as high-quality photos and increased diagnostic accuracy [17], [22]. There are numerous benefits and few limitations when using Ultrasound. Acquiring the device is a very inexpensive process, particularly when compared to other instruments used on the same organs, such as MRI or Positron Emission Tomography (PET). Unlike other acquisition equipment for comparable purposes, Ultrasound scanners are portable and long-lasting. The second major benefit is that, unlike MRI and CT, the Ultrasound machine does not pose any health concerns for pregnant women as the produced signals are entirely safe for both the mother and the fetus [23]. Although the literature exploring the benefits of utilizing AI. technologies for Ultrasound imaging diagnostics in prenatal care has grown, more research is needed to understand the different roles A.I. can have in diagnostic prenatal care using ultrasound images.

### 1.2 Research Objective

Previous reviews on Ultrasound medical images for pregnant women using AI techniques has not been thoroughly conducted. Early reviews have attempted to summarize the use of AI in the prenatal period [22], [24]–[32]. However, these reviews are not comprehensive because they only target specific issues such as fetal cardiac function [24], Down syndrome [25], and the fetal central nervous system [26]. Similarly, other reviews have not focused on AI techniques and Ultrasound image as the primary intervention [27]–[29]. Two reviews covered the use of general AI techniques utilized in the prenatal period through Ultrasound. However, they did not focus on fetal as the primary population in their review [22], [30]. Another systematic review [31] addressed AI classification technologies for fetal health, but it only targeted cardiotocography disease. Other review [32] they introduced various areas of medical AI research including obstetrics but the main focus on US imaging in general. Therefore, a comprehensive survey is necessary to understand the role of AI technologies for the entire prenatal care period using Ultrasound images. This work will be the first comprehensive study that systematically reviews how AI techniques can assist during pregnancy screenings of fetal development and improve fetus growth monitoring. In this survey we aim to (a) Describe the application of AI techniques on different fetal spatial ultrasound dimensions (b) Discuss how different methodologies are used; classification, segmentation, object-detection, regression, and reinforcement learning, (c) Highlight the dataset acquisition and availability (d) Identify practical and research Implications and gaps for future research.

## 2. Methods

### 2.1 Overview

We conducted a survey using method of systematic reviews as reported by Joanna Briggs Institute for systematic reviews [33]. The findings are reported in accordance with the Preferred Reporting Items for Systematic Reviews and Meta-Analyses (PRISMA) [34].

### 2.2 Search strategy

The bibliographic databases used in this study were PubMed, EMBASE, PsycINFO, IEEE Xplore, ACM Digital Library, ScienceDirect, and Web of Science. Google Scholar was also used as a search engine. The first 200 results were filtered from Google Scholar, which returned a significant number of papers sorted by relevancy to the search subject (20 pages). On June 22, 2021, the search began and ended on June 23, 2021. The original query was broad: we searched terms in all fields in all databases; due to the huge number of irrelevant studies while searching in Web of Science and ScienceDirect, we only searched article title, abstract, and keywords. We derived our preliminary search terms from previous works [22], [30] and identified additional terms in consultation with one digital health expert. Multimedia Appendix 1 lists the search strings used to search each bibliographic database.

### 2.3 Study Eligibility Criteria

Studies published in the last eleven years reporting on ‘fetus’ or ‘fetal’ were included in this survey. The population of interest was pregnant women in the first, second, and third trimesters with a specific focus on fetal development during these time periods. We included computer vision Artificial Intelligence interventions that screened the fetal development process during the prenatal period. Ultrasound image was the only medical image addressed in this survey with different synonyms (e.g., Sonography, Ultrasonography, Echography, and Echocardiography). Inclusion and exclusion criteria are clearly defined in Table.1.

**Table 1.** Inclusion and exclusion criteria

| Criteria | Specified Criteria |
| --- | --- |
| Inclusion | • Studies that address fetus/fetal health during pregnancy that are involved AI and US images.<br>• Studies published in English.<br>• Peer-reviewed studies, conference proceedings, and book chapters.<br>• Studies introducing new AI technology that is used for medical image processing.<br>• Proposed algorithms that mainly perform the following tasks; classification, segmentation, object detection, regression, or reinforcement learning.<br>• Studies conducted from 2010 to 2021. |
| Exclusion | • Studies that were out of the scope of fetal health.<br>• Studies that do not use ultrasound as the source medical images such as MRI or CT.<br>• Studies that relied on biometric ultrasound data instead of the images themselves.<br>• Studies reported in a language other than English.<br>• Conference abstracts, reviews magazines and newspapers.<br>• Studies that use traditional image processing technologies without AI |

### 2.4 Study Selection

A web-based systematic review tool called Rayyan Software aided the research selection process and sped up the screening step [35]. The studies were chosen using a two-step process. First, all the retrieved papers' titles and abstracts were independently checked by two authors. Both authors independently read the entire texts of the studies considered in the first phase. Any disagreement in the screening process was resolved through consensus by the authors involved.

### 2.5 Data Extraction and Synthesis

To undertake a systematic data extraction, we designed a data extraction form and piloted it using three included studies (Multimedia Appendix 2). Data from the included studies was extracted separately by two authors. Discrepancies were addressed by consultation with the other authors. The data was synthesized using a narrative method and the results of the research were categorized into groups based on Fetal organ and AI technique tasks.

## 3. Results

### 3.1 Search Results

As shown in Figure 1, the search yielded a total of 1269 citations. After excluding 457 duplicates, there were 812 unique titles and abstracts. A total of 598 citations were excluded after evaluating the titles and abstracts.After full-text screening, n= 107 citations were excluded from the remaining n= 214 papers. The narrative synthesis includes a total of 107 studies. Across screening steps, both authors reported the excluding reason for each study to ensure reliability of the work. The total excluded studies were n= 704 articles. These studies did not meet selection criteria due to the following reasons: (1) irrelevant, (2) wrong intervention, (3) wrong population, (4) wrong publication type, (5) unavailable, and (6) foreign language article. These terminologies are defined in Table 2.

**Table 2.** Definition of the excluded terminologies

| Terminologies | Definition |
| --- | --- |
| Irrelevant | • Publication that are not related to the scope of this survey |
| Wrong intervention | • Publication that targets fetus health but not using AI technology and Ultrasound image |
| Wrong population | • Publication that use AI technology and Ultrasound image but did not target fetus health |
| Wrong publication type | • Publication that is conference abstract, review, magazine, and newspaper |
| Unavailable | • Publication that is not accessible or cannot be found |
| Foreign language | • Publication that is not written in English language |

### 3.2 Assessment of Bibliometrics Analysis

To validate the selected studies, we used a bibliometrics analysis [36] tool that explores the literature through citations and automatically recommends highly related articles to our scope. This ensured that we were unlikely to miss any relevant study. Figure 2. Provides a citations map between selected studies over time. The map shows that all selected studies are relevant over time from 2010 to 2021and recommends nine studies that may be relevant to our scope based on the interactive citation. However, these studies were excluded because they did not meet our inclusion criteria. Therefore, we concluded that our search was comprehensive and included most of the relevant studies.

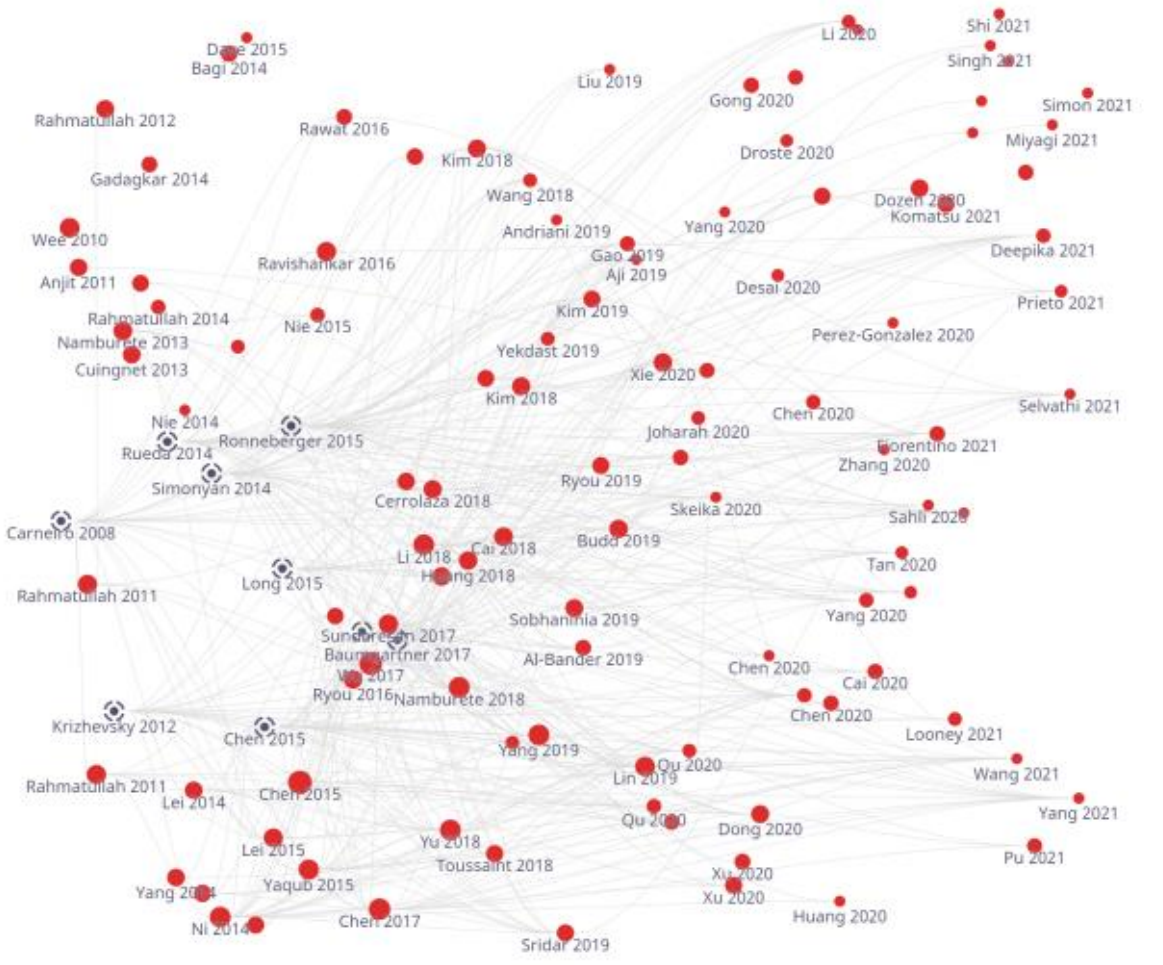

**Figure 2**. Literature Map showing the selected studies in red dot and recommended studies in black dot.

## 3.3 Identification of result themes

In Figure 1, we grouped the studies by the fetus organ that each study addresses including: 1) fetus head (n=43, 40.18%), 2) fetus body (n=31,28.97%), 3) fetus heart (n=13,12.14%), 4) fetus face (n=10, 9.34%), and 5) fetus abdomen (n=10, 9.34%). Each group is classified by sub-group as detailed in Figure 3. In addition, most of the studies used 2D US (n=88, 82.24%) and 3D/4D US were rare and reported only in (n=19, 17.75%). AI tasks are grouped based on the most common use. We found that classification was used in (n=42, 39.25%), segmentation was used in (n=31, 28.97%), both classification and segmentation were used together in (n=16, 14.95%), object-detection, regression, and reinforcement learning were seen as miscellaneous in (n=18, 16.82%).

## 3.4 Definition of result themes

### 3.4.1 Ultrasound imaging modalities

There was a total of (n=88, 82.24%) studies that utilized 2D US images. therefore, here is a brief about what's the 2D Ultrasound. Sound waves are used in all ultrasounds to produce an image. The conventional ultrasound picture of a growing fetus is a 2D image. The fetus's internal organs may be seen with a 2D ultrasound, which provides outlines and flat-looking pictures. 2D ultrasounds have been widely available for a long time and have a very good safety record. These devices do not pose the same dangers as X-rays because they employ non-ionizing rather than ionizing radiation [22]. Ultrasounds are typically conducted at least once throughout pregnancy, most often between 18 and 22 weeks in the second trimester [37]. This examination, also known as a level two ultrasound or anatomy scan, is performed to monitor fetus's development. Ultrasounds may be used to examine a variety of things during pregnancy, including: 1) how the fetus is developing; 2) the gestational age of the fetus; 3) any abnormalities within the uterus, ovaries, cervix, or placenta; 4) The number of fetuses you're carrying; 5) Any difficulties the fetus may be experiencing; 6) the fetus's heart rate; 7) fetal growth and location in the uterus, 8) amniotic fluid level, 9) sex assignment, 10) signs of congenital defects, and 11) signs of Down syndrome [38], [39].

There was a total of (n=18, 16.8%) studies that utilized 3D. Therefore, here is a brief about what's the 3D Ultrasound. Observing three-dimensional (3D) ultrasound images has increasingly grown in recent years. Although 3D ultrasounds may be beneficial in identifying facial or skeletal abnormalities [40], 2D ultrasounds are commonly utilized in medical settings because they can clearly reveal the interior organs of a growing fetus. The picture of a 3D ultrasound is created by putting together several 2D photos obtained from various angles. Many parents like 3D photos because they believe they can see their baby's face more clearly than they can with flat 2D images.

Despite this, the Food and Drug Administration (FDA) does not recommend using a 3D ultrasound for entertainment purposes [41]. The "as low as reasonably attainable" (ALARA) approach directs ultrasound technologists to be used solely in a clinical setting in order to reduce exposure to heat and radiation [42]. While ultrasound is generally thought to be harmless, there is insufficient data to determine what long-term exposure to ultrasound may do to a fetus or a pregnant woman. There is no way of knowing how long a session will take or if the ultrasound equipment will work correctly in non-clinical situations such as places that give "keepsake" photos [43].

4D ultrasound is identical to a 3D ultrasound, but the image it creates is updated constantly – comparable to a moving image. This sort of ultrasound is usually performed for fun rather than for medical concerns. As previously mentioned, because ultrasound is medical equipment that should only be used for medical purposes [44], the FDA does not advocate getting ultrasounds for enjoyment or bonding purposes. Unless their doctor or midwife has recommended it as part of prenatal care, pregnant patients should avoid non-medical locations that provide ultrasounds[43]. Though there's no proven risk but both 3D and 4D ultrasounds employ higher-than-normal amounts of ultrasound energy and sessions are longer compared to 2D which may have fetal adverse effects[45], [46].

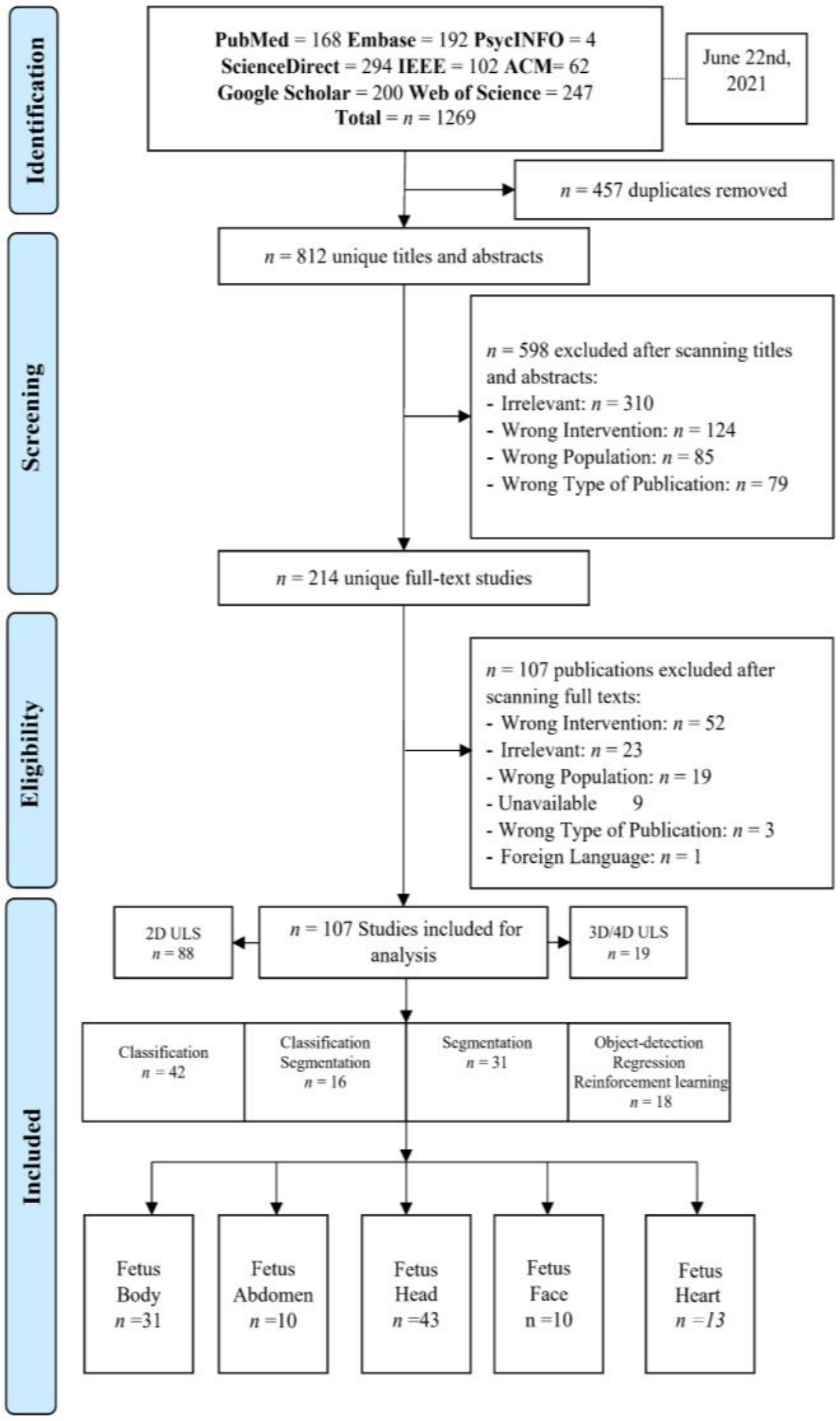

**Figure 1**. PRISMA diagram showing our literature search inclusion process.

### 3.4.2 AI image processing task

There was a total of (n=42, 39.25%) studies that utilized classification, which is the process of assigning one or more labels to an image, and it is one of the most fundamental problems concerning accurate computer vision and pattern recognition. It has many applications, including image and video retrieval, video surveillance, web content analysis, human-computer interaction, and biometrics. Feature coding is an important part of image classification, and several coding methods have been developed in recent years. In general, image classification entails extracting picture characteristics before classifying them. As a result, the main aspect of utilizing image classification is understanding how to extract and evaluate image characteristics [47].

There was a total of (n=31, 28.97%) studies that utilized segmentation, which one of the most complex task in medical image processing, that distinguishing the pixels of organs or lesions from background medical images such as CT or MRI scans. A number of researchers have suggested different automatic segmentation methods by using existing technology[48]. Previously, traditional techniques like edge detection filters and mathematical algorithms were used to construct earlier systems [49]. After this, machine learning techniques for extracting hand-crafted characteristics were the dominating method for a substantial time period. The main issue for creating such a system has always been designing and extracting these characteristics, and the complexity of these methods have been seen as a substantial barrier to deployment. In the 2000's, deep learning methods began to grow in popularity because of advancements in technology as this method had significantly better skills in image processing jobs. Deep learning methods have emerged as a top choice for image segmentation, particularly medical image segmentation, due to their promising capabilities [50].

There was a total of (n=12, 11.21%) studies that utilized object detection techniques, which is the process of locating and classifying items. A detection method is used in biomedical images to determine the regions where the patient's lesions are situated as box coordinates. There are two kinds of deep learning-based object detection. These region proposal-based algorithms are one example. Using a selective search technique, this method extracts different kinds of patches from input images. After that, the trained model determines if each patch contains numerous items and classifies them according to their region of interest (ROI). In particular, the region proposal network, was created to speed up the detection process.

Object identification in the other methods is done using the regression-based algorithm as a one-stage network. These methods use image pixels to directly identify and detect bounding box coordinates and class probabilities within entire images [6].

There was a total of (n=2, 1.8%) studies that utilized Reinforcement Learning (RL), the concept behind RL is that an artificial agent learns by interacting with its surroundings. It enables agents to autonomously identify the best behavior to exhibit in each situation to optimize performance on specified metrics. The basic concept underlying reinforcement learning is made up of many components. The RL agent is the process's decision-maker, and it tries to perform an action that has been recognized by the environment. Depending on the activity performed, the agent receives a reward or penalty from its surroundings. Exploration and exploitation are used by RL agents to determine which activities provide the most rewards. In addition, the agent obtains information about the condition of the environment [51]. For medical image analysis applications, RL provides a powerful framework. RL has been effectively utilized in a variety of applications, including landmark localization, object detection, and registration using image-based parameter inference. RL has also been shown to be a viable option for handling complex optimization problems such as parameter tuning, augmentation strategy selection, and neural architecture search. Existing RL applications for medical imaging can be split into three categories: parametric medical image analysis, medical image optimization, and numerous miscellaneous applications [52].

### 3.5 Description of Included Studies

As shown in Table 3, more than half of the studies (n= 53, 49.53%) were obtained from journal articles, while the other half were found in conference proceedings (n= 43, 40.18%) and book chapters (n=11, 10.28%). Studies were published between 2010 and 2021. The majority of studies were published in 2020 (n=30, 28.03%), followed by 2021 (n=19, 17.75%), 2019 (n=14, 13.08%), 2018 (n=11, 10.28), 2014 (n=9, 8.41%), 2017 (n=6, 5.60%), 2015 (n=6, 5.60%), 2016 (n=4, 3.73%), 2013 (n=3, 2.80%), 2011 (n=3, 2.80%), 2012 (n=1, 0.93%), and 2010 (n=1, 0.93%). In regard to studies examining fetal organs, fetus skull localization and measurement was reported on in (n=25, 23.36%) studies, followed by fetal part structures (n=13, 12.14%) and brain standard plane in (n=13, 12.14%) studies. Abdominal anatomical landmarks were reported in (n=10, 9.34%) studies. Additionally, fetus body anatomical structures were reported in (n=8, 7.47%) studies. Heart disease was reported on in (n=7, 6.54%) studies. Growth disease (n=6, 5.60%) and brain disease were also reported in (n=5, 4.67%) studies. The view of heart chambers was reported on in (n=6, 5.60%) studies. Fetal facial standard planes were reported in (n=5, 4.67%) studies. Gestational age (n=3, 2.80%) and face anatomical landmarks were reported on in (n=3, 2.80%) studies. Facial expressions were reported on in (n=2, 1.86%) studies, and gender identification was reported on only in (n=1, 0.93%) study. In addition, most of the studies were conducted in China (n=41, 38.31%), followed by the UK (n=25, 23.36%), and India (n=14, 13.08%). Surprisingly, some institutes mainly focused on this field and contributed high number of studies. For example, many (n=20, 18.69%) studies were conducted at the University of Oxford, UK and several (n=13, 12.14%) were conducted in Shenzhen University, China.

### 3.6 Organization of the survey findings

The survey is organized as shown in Figure 3. We divided AI for Fetal US medical image into five parts (Section 4), which classify fetus research by organs. In the US image, each group was divided into sub-sections based on recognized characteristics or target illnesses. This survey discusses how different tasks are used; classification, segmentation, object-detection, regression, and reinforcement learning. We discuss the leverage of public dataset in Section 5 and compared similar work depending on the fetal organ, as shown in Section 4. We also divided the discussion (Section 6) into two subsections: principle findings, and practical and scientific implications. Lastly, we highlighted research gaps and future work in Section 7.

**Table 3.** Characteristics of the included studies (n=107).

| **Characteristics** | Number of studies |
|---|---|
| **Type of publication** | (n, %) |
| Journal article | **(53, 49.53)** |
| Conference proceedings | **(43, 40.18)** |
| Book chapter | **(11, 10.28)** |
| **Year of publication** | |
| 2021 | **(19, 17.75)** |
| 2020 | **(30, 28.03)** |
| 2019 | **(14, 13.08)** |
| 2018 | **(11, 10.28)** |
| 2017 | **(6, 5.60)** |
| 2016 | **(4, 3.73)** |
| 2015 | **(6, 5.60)** |
| 2014 | **(9, 8.41)** |
| 2013 | **(3, 2.80)** |
| 2012 | **(1, 0.93)** |
| 2011 | **(3, 2.80)** |
| 2010 | **(1, 0.93)** |
| **Fetal Organ** | |
| Fetus Body | **(31, 28.97)** |
| Fetal part structures | (13, 12.14) |
| Anatomical structures | (8, 7.47) |
| Growth disease | (6, 5.60) |
| Gestational age | (3, 2.80) |
| Gender | (1, 0.93) |
| Fetus Head | **(43, 40.18)** |
| Skull localization and measurement | (25, 23.36) |
| Brain standard plane | (13, 12.14) |
| Brain disease | (5, 4.67) |
| Fetus Face | **(10, 9.34)** |
| Fetal facial standard planes | (5, 4.67) |
| Face anatomical landmarks | (3, 2.80) |
| Facial expressions | (2, 1.86) |
| Fetus Heart | **(13,12.14)** |
| Heart disease | (7, 6.54) |
| Heart chambers view | (6, 5.60) |
| Fetus Abdomen | **(10, 9.34)** |
| Abdominal anatomical landmarks | (10, 9.34) |
| **Publication Country and top institute** | |
| China | **(41, 38.31)** |
| Shenzhen University | (13, 12.14) |
| Beihang University | (5, 4.67) |
| Chinese University of Hong Kong | (4, 3.73) |
| Hebei University of Technology | (2, 1.86) |
| Fudan University | (2, 1.86) |
| Shanghai Jiao Tong University | (2, 1.86) |
| South China University of Technology | (2, 1.86) |
| Other institutes | (11, 10.28) |
| UK | **(25, 23.36)** |
| University of Oxford | (20, 18.69) |
| Imperial College London | (4, 3.73) |
| King's College, London | (1, 0.93) |
| India | **(14, 13.08)** |
| Japan | **(4, 3.73)** |
| Indonesia | **(4, 3.73)** |
| USA | **(3, 2.80)** |
| South Korea | **(3, 2.80)** |
| Iran | **(3, 2.80)** |
| Australia | **(1, 0.93)** |
| Canada | **(1, 0.93)** |
| Mexico | **(1, 0.93)** |
| France | **(1, 0.93)** |
| Italy | **(1, 0.93)** |
| Tunisia | **(1, 0.93)** |
| Spain | **(1, 0.93)** |
| Iraq | **(1, 0.93)** |
| Brazil | **(1, 0.93)** |
| Malaysia | **(1, 0.93)** |

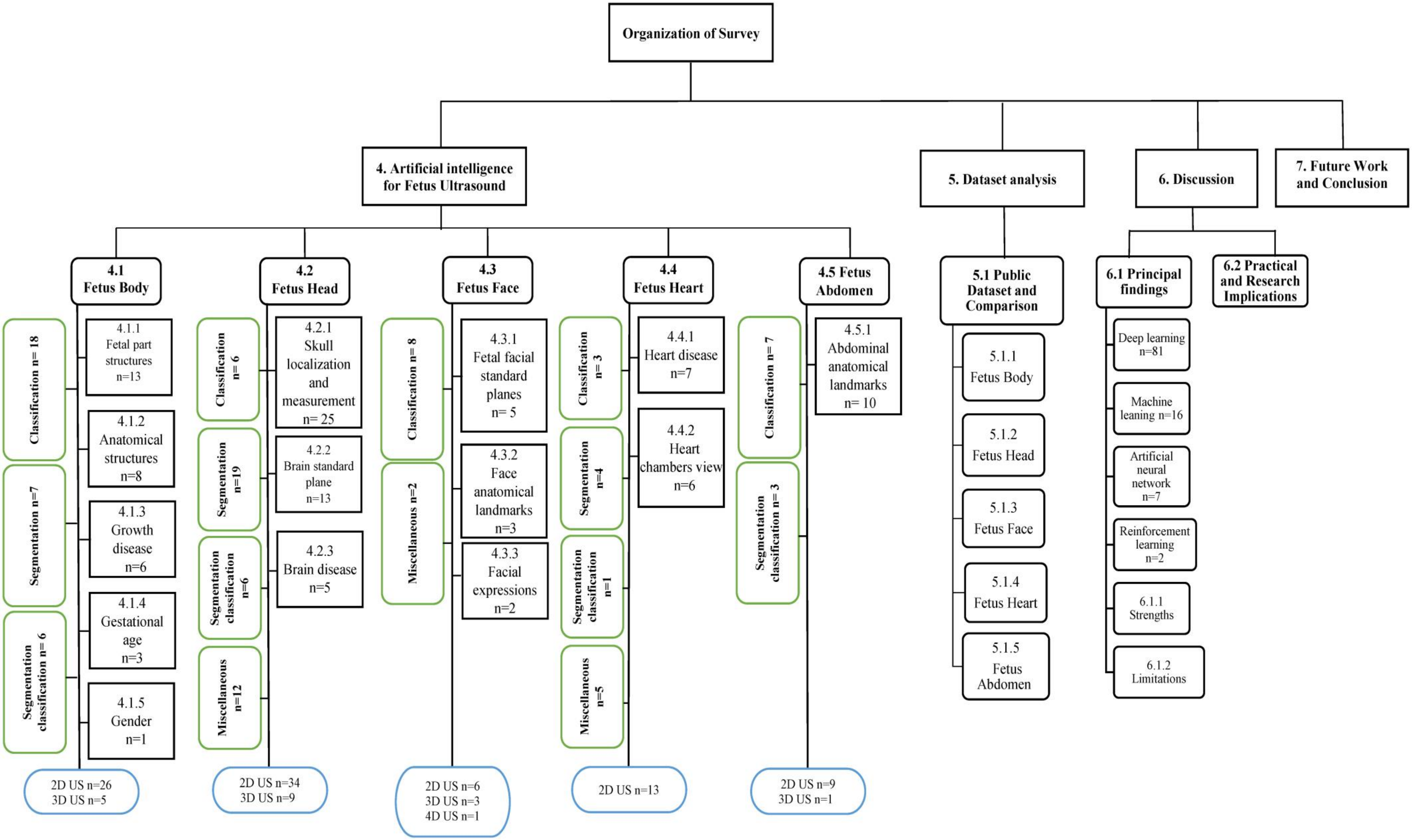

**Figure 3**. Organization of this survey paper

## 4. Artificial intelligence for fetus Ultrasound

### 4.1 Fetus Body

As shown in Table 3, the overall purpose of several (n= 31, 28.97%) studies was to identify general characters of the fetus (e.g., gestational age, gender) or diseases such as Intrauterine Growth Restriction (IUGR). In addition, most of the studies aimed to identify the fetus itself in the uterus or fetus part structure during different trimesters. The first trimester is from week 1 to the end of week 12. The second trimester is from week 13 to the end of week 26, and the third trimester is from week 27 to the end of the pregnancy. We also report on a performance comparison between the various techniques for each fetus body group including objective, backbone methods, optimization, fetal age, best obtained result, and observations (see Table 5).

#### 4.1.1 Fetal Part Structures

- **Classification**

For identification and localization of fetal part structure, image classification was widely used in (n= 9, 8.41%) studies. Only 2D US images were utilized for classification purposes. Studies [53]–[55] used a classification task to identify and locate fetal skull, heart, and abdomen from 2D US images in the second trimester. Besides that, in [56], classification was used to locate the exact abdominal circumference plane (ACP), and in [57], more results were obtained by locating head circumference plane (HCP), the abdominal circumference plane (ACP), and the femur length plane (FLP). Furthermore, in studies [58]–[60] researching the second and third trimester, multi-classes were used to identify and locate various parts including fetal abdominal standard plane (FASP); 2) fetal face axial standard plane (FFASP); and 3) fetal four-chamber view standard plane (FFVSP) of heart. Study [61] located the mother's cervix in addition to abdomen, brain, femur, and thorax.

- **Segmentation**

Image segmentation task was used in (n=3, 2.80%) studies for the purpose of fetal part structure identification. 2D US imaging [62] was used to segment neonatal hip bone including seven key structures: 1) chondro-osseous border (CB), 2) femoral head (FH), 3) synovial fold (SF), 4) joint capsule & perichondrium (JCP), 5) labrum (La), 6) cartilaginous roof (CR), and 7) bony roof (BR). Only one study in [63] provided segmentation model that able to segment fetus kidney using 3D US in the third trimester. Segmentation was used to locate fetal head, femur, and humerus using 2D US as seen in [64].

- **Classification and segmentation**

In [65], whole fetal segmentation followed by classification tasks were used to locate the head, abdomen (in sagittal view), and limbs (in axial view) using 3D US images taken in the first trimester.

#### 4.1.2 Anatomical Part Structures

- **Classification**

For identification and localization of anatomical structures, image classification was used in (n= 4, 3.73%) studies. Only one study [66] used 3D US images and other studies used 2D US [67]–[69]. In [66], the whole fetus was localized in the sagittal plane and classifier was then applied to the axial images to localize one of three classes (head, body and non-fetal) during the second trimester. Multi-classification methods were utilized in [67] to localize head, spine, thorax, abdomen, limbs, and placenta in the third trimester. Moreover, binary classification was used in [68] to identify abdomen versus non-abdomen. In [69], classification method was used to detection and measurement nuchal translucency (NT) in the beginning of second trimester. Moreover, monitoring NT combined with maternal age can provide effective insight of screening Down syndrome.

- **Segmentation**

An image segmentation task was used in (n= 4, 3.73%) studies for the purpose of fetal mage segmentation was target task anatomical structures identification. In [70], [71], 3D US were utilized to segment fetus, gestational sac amniotic fluid, and placenta in the beginning of second trimester. However, in [72], 2D US was used to segment amniotic fluid and the fetus in the late trimester. Finally, 2D US in [73] was utilized to segment and measure nuchal translucency (NT) in the first trimester of pregnancy.

#### 4.1.3 Growth disease diagnosis

- **Classification**

For diagnosis of fetal growth, disease image classification was conducted in (n= 4, 3.73%) studies using 2D US. In [74] binary classification was used for early diagnosis of Intrauterine Growth Restriction (IUGR) at the third trimester of pregnancy. The features considered to determine a diagnosis of IUGR are gestational age (GA), bi-parietal diameter (BPD), abdominal circumference (AC), head circumference (HC), and femur length (FL). Additionally, binary-classification was used in the second trimester of pregnancy to identify normal vs abnormal fetus growth as seen in [75]. In [76], binary-classification is used to identify the normal growth of the ovum by distinguishing Blighted Ovum from healthy Ovum. Lastly, the binary classification was used to distinguish between healthy fetuses and those with Down syndrome [77].

- **Classification and segmentation**

For diagnosis of fetal growth disease, image classification along with segmentation were used in (n= 2, 1.86%) studies using 2D US. In [78], [79], segmentation of the region of interest (ROI), followed by classification to diagnosis (IUGR) (normal vs abnormal) were carried out. This was done by using US Images taken in both second and third trimester of pregnancy. This classification relied on the measurement of the following: 1) Fetal abdominal circumference (AC), 2) Head circumference (HC), 3) Biparietal diameter (BPD), 4) Femur length.

#### 4.1.4 Gestational age (GA) estimation

- **Classification and segmentation**

Both classification as well as segmentation tasks were used to estimate GA in (n=3, 2.80%) studies using 2D US taken in the second and third trimester of pregnancy. In [80], the trans-cerebellar diameter (TCD) measurement was used to estimate GA in week. The TC plane frames are extracted from the US images using classification. Segmentation then localizes the TC structure and performs automated TCD estimation, from which the GA can thereafter be estimated via an equation. In [81], 2D US was used to estimate GA based on region of fetal lung in the second and third trimester. In the first stage, the segmentation is to learn the recognition of fetal lung region in the ultrasound images. Classification is also used to accurately estimate the gestational age from the fetal lung region of ultrasound images. Several fetal structures were used to estimate GA in [82], which focused on the second trimester. In this study, 2D US images were classified into four categories: head (BPD and HC), abdomen (AC), femur (FL), and fetus (crown-rump length: CRL). Then, the regions of interest (i.e., head, abdomen, and femur) were segmented and the results of biometry measurements were used to estimate GA.

#### 4.1.5 Gender identification

- **Classification**

Binary-classification was used in [83] to identify the gender of the fetus. Image preprocessing, image segmentation, and feature extraction (shape description) were used to obtain the value of the feature extraction and gender classification. This task is categorized as classification because a segmentation task is not clearly defined.

## 4.2 Fetus Head

As shown in Table 3, the primary purpose of (n= 43, 40.18%) studies is to identify and localize fetus skull (e.g., head circumference), brain standard planes (e.g., Lateral sulcus (LS), thalamus (T), choroid plexus (CP), cavum septi pellucidi (CSP)) or brain diseases (e.g., hydrocephalus (premature GA), ventriculomegaly, central nervous system (CNS) malformations)). The following subsection discusses each category based on the implemented task. Table 6 presents comparison between the various techniques for each fetus head group including objective, backbone methods, optimization, fetal age, best obtained result, and observations. studies for each group under the fetus head category and provides an overview of methodology and observations.

### 4.2.1 Skull localization and measurement

- **Classification**

Classification tasks for skull localization and head circumference are rarely used, as seen only in (n=3, 2.80%) studies. Studies [84], [85] used classification to localize the region of interest (ROI) and identify the fetus head based on 2D US taken in first and second trimester respectively. 3D US taken in first trimester were used to detect a fetal head in one study [86].

- **Segmentation**

Skull localization and head circumference (HC) in most of the studies (n=16, 14.81%) used segmentation task. 2D US was used in 16 studies and one study used 3D US. Various network architecture were used to segment and locate skull, then perform head circumference as seen in [87]–[99]. Besides identifying HC, in [100]segmentation was also used to find fetal biparietal diameter (BPD). Further investigation shows that the work completed in [101] was the first investigation about whole fetal head segmentation in 3D US. The segmentation network for skull localization in [91] was tested to identify a view of the four heart chambers on different datasets.

- **Miscellaneous**

Along with segmentation, another method was used to locate and identify the fetus skull in (n=6, 5.60%) studies. As seen in [102], regression was used along with segmentation to identify HC, BPD, and occipitofrontal diameter (OFD). Besides this, in [103], neural network was used to train the saliency predictor and predict saliency maps to measure HC and identify the alignment of both trans ventricular (TV) as well as trans cerebellar (TC). In addition, an object detection task was used along with segmentation in [104] to locate the fetal head at the end of the first trimester. In [105], object detection was used for head localization and centering and regression to delineate the HC accurately. In [106], classification and segmentation tasks were used by utilizing a multi-Task network to identify head composition.

### 4.2.2 Brain standard plane

- **Classification**

The classification was used in (n=2, 1.86%) studies to identify the brain standard plane, and both studies used 2D US images taken in the second and third trimesters. In [107], [108], two different classification network architectures were used to identify six fetal brain standard planes. These architectures include horizontal transverse section of the thalamus, horizontal transverse section of the lateral ventricle, transverse section of the cerebellum, midsagittal plane, paracentral sagittal section, and coronal section of the anterior horn of the lateral ventricles.

- **Segmentation**

The segmentation task was used in (n=3, 2.80%) studies to identify the brain standard plane. Each of these studies used 2D US images. In [109], 2D US images taken in the second trimester were used to segment fetal cerebellum structure. In addition, in [110], 2D US images taken in the third trimester were used to segment fetal Middle Cerebral Artery (MCA). Authors in [111] utilized 3D US images in the second trimester to formulate the segmentation as a classification problem to identify the following brain planes; background, Choroid Plexus (CP), Lateral Posterior Ventricle Cavity (PVC), the Cavum Septum Pellucidi (CSP), and Cerebellum (CER).

- **Classification and segmentation**

Classification with segmentation is employed in (n=3, 2.80%) studies to identify the brain standard plane. These studies all used 3D US images. In [112], both tasks were used to identify brain alignment based on skull boundaries and then head segmentation, eye localization, and prediction of brain orientation in the second and third trimesters. Moreover, in [113], segmentation followed by a classification task was used to detect cavum septi pellucidi (CSP), thalami (Tha), lateral ventricles (LV), cerebellum (CE), and cisterna magna (CM) in the second trimester. Segmentation and classification are employed by 3D US in [114] to identify the skull, midsagittal plane, and orbits of the eyes in the second trimester of pregnancy.

- **Miscellaneous**

For brain standard planes identification in (n=5, 4.62%), methods different than previously mentioned were used, including; segmentation with object detection, object detection, classification with object detection, and Reinforcement Learning (RL). As seen in [115], the RL-based technique was employed for the first time on 3D US images to localize standard brain planes, including trans-thalamic (TT) and trans-cerebellar (TC) in the second and third trimesters. RL is used in [116] to localize the following: mid-sagittal (S), transverse (T), and coronal (C) planes in volumes and trans-thalamic (TT), trans-ventricular (TV), and trans-cerebellar (TC)). Object detection architecture is utilized on 2D US images in [117] to localize the trans-thalamic plane in the second trimester of pregnancy, including lateral sulcus (LS), thalamus (T), choroid plexus (CP), cavum septi pellucid (CSP), and third ventricle (TV). In [118], classification with object detection was used to detect Lateral sulcus (LS), thalamus (T), choroid plexus (CP), cavum septi pellucidi (CSP), third ventricle (TV), brain midline (BM). In contrast,[119]used classification with object detection to localize cavum septum pellucidum (CSP) and ambient cistern (AC and cerebellum), as well as to measure head circumference (HC) and biparietal diameter (BPD).

### 4.2.3 Brain disease

- **Classification**

The Binary-classification technique utilized 2D US images taken in the second trimester as seen in [120] to detect hydrocephalus disease using premature GA.

- **Classification and segmentation**

Studies [121], [122] both used 2D US images taken in the second and third trimesters to segment craniocerebral and identify abnormalities or specific diseases. As seen in [121], central nervous system (CNS) malformations were detected using binary classification. Moreover, in [122], multiclassification was used to identify the following problems: TV planes contained occurrences of ventriculomegaly and hydrocephalus, and TC planes contained occurrences of Blake pouch cyst (BPC).

- **Miscellaneous**

In [123], [124] different methods were used to detect ventriculomegaly disease based on 2D US images taken in the second trimester. In study [123], object detection and regression first identify fetal brain ultrasound images from standard axial planes as normal or abnormal. Second regression was then used to find the lateral ventricular regions of images with big lateral ventricular width. Then, the width was anticipated with a modest error based on these regions. Furthermore, [124] was the first study to propose an object-detection-based automatic measurement approach for fetal lateral ventricles (LVs) based on 2D US images. The approach can both distinguish and locate the fetal LV automatically as well as measure the LV's scope quickly and accurately.

## 4.3 Fetus Face

As shown in Table 3, the primary purpose of (n= 10, 9.34%) studies was to identify and localize fetus face features such as the fetal facial standard plane (FFSP) (i.e., axial, coronal, and sagittal plane), face anatomical landmarks (e.g., Nasal Bone), and facial expressions (i.e., sad, happy). The following subsection discusses each category based on the implemented task. Table 7 presents comparison between the various techniques for each fetus face group including objective, backbone methods, optimization, fetal age, best obtained result, and observations.

### 4.3.1 Fetal facial standard plane (FFSP)

- **Classification**

Classification was used to classify the FFSP using 2D US Images as seen in (n=5, 4.67%). Using 2D US images that taken in the second and third trimesters, four studies [125]–[128] identify ocular axial planes (OAP), the median sagittal planes (MSP), and the nasolabial coronal planes (NCP). In addition, authors in [129] were able to identify the FFSP using 2D US images taken in the second trimester.

### 4.3.2 Face anatomical landmarks

- **Miscellaneous**

Different methods were used to identify face anatomical landmarks in (n=3, 2.80) studies. In [130], 3D US images taken in the second and third trimesters were segmented to identify background, face mask (excluding facial structures), eyes, nose, and lips. In another study the object detection method was used on 3D US images in [131] to detect the left fetal eye, middle eyebrow, right eye, nose, and chin. Furthermore, classification used 2D US images taken in the first and second trimesters to detect the nasal bone. This was done to enhance the detection rate of Down syndrome, as seen in [132].

### 4.3.3 Facial expressions

- **Classification**

For the first time, 4D US images were utilized using the multi-classification method to identify fetus facial expression into Sad, Normal, and Happy as seen in [133]. Study [134] used 2D US images in the second and third trimesters to classify fetal facial expression into eye blinking, mouthing without any expression, scowling, and yawning.

## 4.4 Fetus Heart

As shown in Table 3, the primary purpose of (n= 13, 14.04%) studies is to identify and localize fetus heart diseases and the fetus heart chambers view. The following subsection discusses each category based on the implemented task. Table 8 presents comparison between the various techniques for each fetus heart group including objective, backbone methods, optimization, fetal age, best obtained result, and observations.

### 4.4.1 Heart disease

- **Classification**

Two studies used classification methods to identify heart diseases based on 2D US images taken in the second trimester. In [135], binary-classification task was utilized to identify Down syndrome vs normal fetuses based on identifying the echogenic intracardiac foci (EIF). Furthermore, in [136], binary- classification task was utilized to detect hypoplastic left heart syndrome (HLHS) vs healthy cases based on the four-chamber heart (4CH), left ventricular outflow tract (LVOT), and right ventricular outflow tract (RVOT).

- **Segmentation**

2D US images were utilized by multi-class segmentation to identify heart disease, including: left heart syndrome (HLHS), total anomalous pulmonary venous connection (TAPVC), pulmonary atresia with intact ventricular septum (PA/IVS), endocardial cushion defect (ECD), fetal cardiac rhabdomyoma (FCR), and Ebstein's anomaly (EA) [137].

- **Classification and segmentation**

Classification and segmentation are used to detect congenital heart disease (CHD) based on 2D US images taken in the second trimester [138]. Cases were classified into normal hearts vs CHD. This classification was orchestrated by identifying five views of the heart in fetal CHD screening, including three-vessel trachea (3VT), three-vessel view (3VV), left-ventricular outflow tract (LVOT), axial four chambers (A4C), and abdomen (ABDO).

- **Miscellaneous**

To identify heart disease, various methods were proposed based on the capabilities of 2D US images. Classification with object detection were utilized in [139], [140]. Fetal congenital heart disease (FHD) can be detected in the second and third trimesters based on how quickly the fetus grows between gestational weeks, and how the shape of the Four Chambers heart changes over time [139]. Furthermore, classification with object detection was used to detect cardiac substructures and structural abnormalities in the second and third trimesters [140]. Lastly, segmentation with object detection was used to locate the ventricular septum in 2D US images in the second trimester, as seen in [141].

### 4.4.2 Heart chambers view

- **Classification**

2D US images taken in the second and third trimesters were utilized in [142] to propose a classification task utilized to localize the four chambers (4C), the left ventricular outflow tract (LVOT), the three vessels (3V), and the background (BG).

- **Segmentation**

Segmentation was used in (n=3, 2.80%) studies based on 2D US images. In two studies [143], [144], seven critical anatomical structures in the apical four-chamber (A4C) view were segmented, including: left atrium (LA), right atrium (RA), left ventricle (LV), right ventricle (RV), descending aorta (DAO), epicardium (EP) and thorax. In another study [145], segmentation was used to locate four-chamber views (4C), left ventricular outflow tract view (LVOT), and three-vessel view (3V) in the second and third trimesters.

- **Miscellaneous**

2D US images were used in (n=2, 1.86) studies, and both studies utilized classification with object detection. In the second trimester [146], the first classification conducted was the cardiac four-chamber plane (CFP) into non-CFPs and CFPs (i.e., apical, bottom, and parasternal CFPs). The second task was to then classify the CFPs in terms of the zoom and gain of 2D US images. In addition, object detection was utilized to detect anatomical structures in the CFPs, including left atrial pulmonary vein angle (PVA), apex cordis and moderator band (ACMB), and multiple ribs (MRs). In [147], object detection was used in the second and third trimesters to extract attention regions for improving classification performance and determining the four-chamber view, including detection of end-systolic (ES) and end-diastolic (ED).

## 4.5 Fetus Abdomen

As shown in Table 3, the primary purpose of (n= 10,

10.8%) studies was to identify and localize the fetus's abdomen, including abdominal anatomical landmarks (i.e., stomach bubble (SB), umbilical vein (UV), and spine (SP)). The following subsection discusses this category based on the implemented task. Table 9 presents comparison between the various techniques for each fetus abdomen including objective, backbone methods, optimization, fetal age, best obtained result, and observations.

#### 4.5.1 Abdominal anatomical landmarks

- **Classification**

Classification task was used in (n=7, 6.54%) studies to localize abdominal anatomical landmarks. 2D US images were used in seven studies [148]–[153] and 3D US images in only one study [154]. In [148], [150], [152], [154], various classifiers were utilized to localize stomach bubble (SB) and umbilical vein (UV), in the first and second trimester as seen in [148], [154], and in the second and third trimesters as seen in [150], [152]. Furthermore, works in [149], [151], [153] located the spine (SP) besides SB and UV on the same trimesters.

- **Classification and segmentation**

2D US images taken within random trimesters were used in [155], [156]. In [155], SB, UV, and AF were localized. Further, abdominal circumference (AC), spine position, and bone regions were estimated. In [156], 2D US images were classified into normal vs abnormal fetuses based on the fetus images such as AF, SB, UV, and SA. Study [157] located SB and the portal section from the UV, and observed amniotic fluid (AF) in the second and third trimesters.

## 5. Dataset Analysis

### 5.1 Public dataset

#### 5.1.1 Fetus Body

In this area, ethical and legal concerns presented the most barriers to comprehensive research. Therefore, datasets were not available to the research community. For example, in this research, studies discussed how the fetal body contains five subsections (fetal part structure, anatomical structure, growth disease, gestational age, and gender identification). Out of the available studies (n=31, 28.9%), one dataset was intended to be available online in [55] related to identifying fetal part structure. Unfortunately, this dataset has not yet been released by the author. The only public dataset that was available online for fetal struture classification was released by Burgos-Artizzu et al. [61], the dataset is a total of 12499 2D fetal ultrasound images, including brain 143, Trans-cerebellum 714, Trans-thalamic 1638, Trans-ventricular 597, abdomen 711, cervix 1626, femur 1040, thorax 1718, and 4213 are unclassified fetal image. Acquiring a high volume of the dataset was also challenging in most of the studies; therefore, multi-data augmentation in (n=11, 10.2%) studies [54], [56], [61], [63], [65], [70], [72], [76], [80] has been observed. These augmented images were employed to boost classification and segmentation performance. SimpleITK library [158] was used for augmentation in [63]. Due to limited data sample in addition to augmentation, k-fold cross-validation method was utilized in (n=8, 7.4%) studies [54], [55], [57], [64], [66], [69], [74], [83]. These cross-validation methods are also used to resolve issues such as overfitting and bias in dataset selection. Only one software named ITK-SNAP [159] was reported in [63] to segment structures in 3D images.

#### 5.1.2 Fetus Head

This section contains three subsections (skull localization, brain standard planes, and brain disease). Total studies (n=43, 40.18%), but we only found one online public dataset called HC18 grand challenge [160]. This dataset was used in (n=13, 12.14%) studies [87]–[89], [91], [92], [95]–[98], [102], [104], [105], [161]. This dataset contains 2D US images collected from 551 pregnant women at different pregnancy trimesters. There were 999 images in the training set and 335 for testing; the sonographer manually annotated the HC. Image augmentation was applied in (n=9, 8.41%) studies [87], [89], [92], [96], [98], [102], [104], [105], [161], and cross-validation was employed in [161]. Table 4 highlights the best result achieved despite the challenges of recording the outcomes of the selected studies.

**Table 4.** Compared between studies that utilized the HC18 dataset

| Study | DSC | HD | DF | ADF |
|---|---|---|---|---|
| [87] | 0.926 | 3.53 | 0.94 | 2.39 |
| [88] | N/A | N/A | 14.9% | N/A |
| [95] | 0.973 | 1.58 | N/A | N/A |
| [96] | 0.968 | N/A | N/A | N/A |
| [97] | 0.973 | N/A | N/A | 2.69 |
| [98] | 0.968 | 1.72 | 1.13 | 2.12 |
| [89] | **0.979** | 1.27 | **0.09** | **1.77** |
| [105] | 0.977 | 1.32 | 0.21 | 1.90 |
| [102] | 0.977 | **0.47** | N/A | 2.03 |
| [104] | 0.977 | 1.39 | 1.49 | 2.33 |
| [91] | 0.971 | 3.23 | N/A | N/A |
| [92] | **0.979** | N/A | N/A | N/A |
| [161] | N/A | N/A | N/A | N/A |

DSC: Dice similarity coefficient, ACC: Accuracy, Pre: Precision, HD: Hausdorff distance, DF: Difference, ADF: Absolute Difference, IoU: Intersection over Union, mPA: mean Pixel Accuracy.

#### 5.1.3 Fetus Face

This section contains three subsections (fetal facial standard planes, face anatomical landmarks, and facial expression). Out of these studies (n=10, 9.34%), we did not find any public dataset available online. However, within the private dataset, image augmentation was applied in (n=3, 2.8%) studies [130], [131], [134]. The k-fold cross-validation was employed in [125], [129], [130]. Lastly, 3D Slicer software [162]was used for image annotation as reported in [130].

#### 5.1.4 Fetus Heart

This section contains two subsections (heart diseases and heart chamber view). Out of these studies (n=13, 12.14%), we did not find any public dataset available online. However, within the private dataset, the image augmentation was applied in (n=4, 3.7%) studies [137], [138], [145], [147]. The k-fold cross-validation was employed in [138], [139], [141], [144], [146].

#### 5.1.5 Fetus Abdomen

This section contains one subsection (abdominal anatomical landmarks). Out of these studies (n=10, 9.34%), we did not find any public dataset available online. However, within the private dataset, image augmentation was applied in (n=4, 3.7%) studies [150], [153], [155], [157]. The cross-validation was not reported in any of the studies.

**Table 5.** Articles published using AI to improve fetus body monitoring: Objective, backbone methods, optimization, fetal age best result, and observations

| Study | Objective | Backbone Methods/Framework | Optimization/ Extractor methods | Fetal age | Best Obtained Result | observations |
|---|---|---|---|---|---|---|
| | **Fetal Part Structures** | | | | | |
| [53] | To identify the fetal skull, heart and abdomen from ultrasound images | SVM as the classifier | Gaussian Mixture Model (GMM)<br>Fisher Vector (FV) | 26th week | Acc: 0.9890<br>mAP: 0.9985 | 1) Methodology and result are clear. 2) Dataset acquisition is clear splitting is not. 3) No data augmentation and no cross-validation. 4) Important performance metrics are missing. |
| [62] | To segment the seven key structures of the neonatal hip joint | Neonatal Hip Bone Segmentation Network (NHBSNet) | Feature Extraction Module<br>Enhanced Dual Attention Module (EDAM)<br>Two-Class Feature Fusion Module (2-Class FFM)<br>Coordinate Convolution Output Head (CCOH) | 16 to 25 weeks. | DSC: 0.8785<br>HD: 8.42 mm<br>AHD: 0.32 mm | 1) Multi-mechanisms were used to build the framework; the method is clear<br>2) Data acquisition is precise, but data splitting could be more interpreted<br>3) No data augmentation and no cross-validation method<br>4) Novel work and unique in the target area. |
| [64] | To segment organs head, femur, and humerus in ultrasound images using multilayer super pixel images features | Simple Linear Iterative Clustering (SLIC)<br>Random forest | Unary pixel<br>shape feature<br>image moment | N/A | Head Femur Humerus<br>Acc: 0.969 0.986 0.981<br>F1 : 0.531 0.456 0.357<br>Rec: 0.803 0.758 0.760<br>Spec: 0.972 0.988 0.9831 | 1) Methodology and results are not presented clearly.<br>2) Dataset is small; acquisition, and splitting are not clear.<br>3) No data augmentation. |
| [63] | To automate kidney segmentation using fully convolutional neural networks. | FCNN: U-Net & UNET++ | N/A | 20 to 40 weeks | DSC: 0.81<br>Jaccard index (JI): 0.69<br>HD: 8.96 mm<br>Mean Surface Distance (MSD) 2.04 mm | 1) Methodology and results are confusing but should be interpreted in a better way. 2) Dataset is very small even after augmentation; data acquisition is precise, but data splitting is not clear. 3) No cross-validation was applied.<br>4) First study that target fetus kidney segmentation. |
| [61] | To evaluate the maturity of current Deep Learning classification techniques for their application in a real maternal-fetal clinical environment | CNN DenseNet-169 | N/A | 18 to 40 weeks | Acc: 0.936<br>Error rate: 0.27 | 1. Methodology and results are not clear. 2. Dataset is large, data acquisition and splitting are precise. 3) No cross-validation was applied. 4) No comparison with state-of-the-art methods. 5) Missing signification performance metrics 6) Models have similar performance compared to an expert when classifying standard planes. |
| [57] | To uses the learnt visual attention maps to guide standard plane detection on all three standard biometry planes: ACP, HCP and FLP. | Temporal SonoEyeNet (TSEN)<br>Temporal attention module: Convolutional LSTM<br>Video classification module: Recurrent Neural Networks (RNNs)+ | CNN feature extractor: VGG-16 | N/A | Pre: 0.894<br>Rec: 0.851<br>F1 : 0.871 | Many technologies were integrated to build hybrid framework, the work is very detailed, but two significant points should be considered:<br>1) Dataset distribution and volume was not clear 2) Paper structure and organization does not help the reader to understand |
| [39] | To support first trimester fetal assessment of multiple fetal anatomies including both visualization and the measurements from a single 3-D ultrasound scan | Multi-Task Fully Convolutional Network (FCN)<br>U-Net | N/A | 11 to 14 weeks | Mean Acc: 0.894<br>Intersection over Union (IoU): 0.76 | 1) clinical 3-D datasets was too small for a 3-D FCN implementation even with data augmentation. 2) Other performance metrics were not used such as F1 score and recall |
| [58] | To automatically classify 14 different fetal structures in 2-D fetal ultrasound images by fusing information from both cropped regions of fetal structures and the whole image | support vector machine (SVM)+ Decision fusion | Fine-tuning AlexNet CNN | 18 to 20 weeks | Mean of all structure's classification<br>Acc: 0.97<br>Pre: 0.76 | 1) Large dataset but no cross validation applied<br>2) paper is not organized in solid way<br>3) Role of SVM and Decision fusion were confusing for the reader |
| [59] | To automatic identification of different standard planes from US images | T-RNN framework:<br>LSTM | Features extracted using J-CNN classifier | 18 to 40 weeks | FASP FFASP FFVSP<br>Acc: 0.94 0.71 0.84<br>Pre: 0.94 0.73 0.89<br>Rec: 0.99 0.95 0.93<br>F1: 0.96 0.96 0.96 | 1) largest dataset no cross-validation applied<br>2) incorrect comparison with other work due to different dataset<br>3) T-RNN framework could be explained in a better way, such as point form |
| [56] | To classify abdominal fetal ultrasound video frames into standard AC planes or background. | M-SEN architecture<br>Discriminator CNN | Generator CNN | N/A | Pre: 0.968<br>Rec: 0.962<br>F1 : 0.965 | 1) Dataset acquisition detailed is missing, the dataset small, and no cross-validation is applied 2) Methodology is not clear 3) Novel and effective algorithm that models sonographer visual attention. |
| [54] | To detect of multiple fetal structures in free-hand ultrasound | CNN<br>Attention Gated LSTM | Class Activation Mapping (CAM) | 28 to 40 weeks | mean IOU is above 0.50<br>Skull Average Pre: 0.94<br>Abdomen Average Pre: 0.93<br>Heart Average Pre: 0.85 | 1) Dataset acquisition and splitting is not apparent<br>2) Only precision and IOU as performance metrics |
| [55] | To extract features from regions inside the images where meaningful structures exist. | Guided Random Forests | Probabilistic Boosting Tree (PBT) | 18 to 22 weeks | Mean Acc for all classes: 0.91 | 1) Dataset acquisition and splitting is not apparent<br>2) Only accuracy as performance metrics |
| [60] | To detect standard planes from US videos | T-RNN<br>LSTM (Transferred RNN) | Spatio-Semporal Feature<br>J-CNN | 18 to 40 weeks | FASP FFASP FFVSP<br>Acc: 0.90 0.86 0.86<br>Pre: 0.74 0.63 0.77<br>Rec: 0.74 0.59 0.61<br>F1: 0.74 0.61 0.68 | 1. largest dataset no cross-validation applied<br>2. incorrect comparison with other work due to different dataset<br>3. T-RNN framework could be explained in a better way, such as point form<br>4. Same work in [59] |
| | **Anatomical Structures** | | | | | |
| [70] | To propose the first and fully automatic framework in the field to simultaneously segment fetus, gestational sac and placenta, | 3D FCN + RNN hierarchical deep supervision mechanism (HiDS) | BiLSTM module denoted as FB-nHiDS | 10 to 14 weeks | DSC: 0.890<br>Coefficient Conformity: 0.749<br>Average Distance of Boundaries: 0.925<br>HD of Boundaries: 9.788 mm | 1) Methodology and results are very detailed but could be interpreted in better way.<br>2) Dataset is very small even after augmentation, data acquisition, and splitting are precise.<br>3) No cross-validation was applied |
| [71] | To segment the placenta, amniotic fluid, and fetus. | FCNN | N/A | 11 to 19 weeks | Placenta AF Fetus<br>DSC: 0.82 0.93 0.88<br>HD: 16.22 10.86 16.57<br>Average HD: 0.85 0.13 0.22 | 1) Methodology and results are very detailed, but multi-model experiments confuse the reader.<br>2) Dataset acquisition is clear, but data splitting is not precise.<br>3) No cross-validation was applied, no augmentation |
| [72] | To segment the amniotic fluid and fetal tissues in fetal US images | The encoder-decoder network based on VGG16 | N/A | 22nd weeks | Mean for all categories:<br>Acc: 0.8099<br>IoU: 0.7083 | 1) Methodology is not explained well. 2) Dataset is very large after augmentation; data acquisition clear but splitting is not precise. 3) Cross-validation was applied.<br>4) Different methods were compared. |
| [66] | To localize the fetus and extract the best fetal biometry planes for the head and abdomen from first trimester 3D fetal US images | CNN | Structured Random Forests | 11 to 13 weeks | Mean ACC: 0.769<br>IoU: ≥ 0.50 | 1) Dataset acquisition and splitting is not apparent<br>2) Methodology steps are confusing for the reader |
| [67] | To detect and localize fetal anatomical regions in 2D US images | ResNet18 | Soft Proposal Layer (SP) | 22 to 32 weeks | ACC: 0.912<br>IoU: 0.393 | 1) Dataset splitting is well explained but source and acquisition not clear<br>2) Validation is not applied |
| [68] | To reliably estimate abdominal circumference | CNN + Gradient Boosting Machine (GBM) | Histogram of Oriented Gradient (HoG) | 15 to 40 weeks | DSC: 0.90 | 1) Dataset acquisition and splitting is not apparent. 2) Methodology is not organized well. 3) Validation is not applied. 4) Only DSC as performance metrics. |
| [69] | To detect and recognize the fetal NT based on 2D ultrasound images by using artificial neural network techniques. | Artificial Neural Network (ANN) | Multilayer Perceptron (MLP) Network<br>Bidirectional Iterations Forward Propagations Method (BIFP) | N/A | ACC: 0.933 (56/60)<br>Spec: 0.966<br>Rec : 0.900 | 1) Small dataset, acquisition and splitting is not apparent<br>2) Framework could be explained in a better way, such as point form |
| [73] | To detect NT region | U-Net NT Segmentation<br>PCA NT Thickness Measurement | VGG16 NT Region Detection | 4 to 12 weeks | IoU: 0.71<br>Detection Error (in pixel): 41.54 | 1)Dataset acquisition is not apparent. 2) Methodology very clear and explained in a consistent way. 3) Only IoU and error rate for evaluation |

(Continued)

**Table 5.** Continued

| Study | Objective | Backbone Methods/Framework | Optimization/ Extractor methods | Fetal age | Best Obtained Result | observations |
|---|---|---|---|---|---|---|
| **Growth disease** | | | | | | |
| [74] | To propose the biometric measurement and classification of IUGR, using OpenGL concepts for extracting the feature values and ANN model is designed for diagnosis and classification | ANN<br>Radial Basis Function (RBF) | OpenGL | various | ACC: 0.94<br>Root Mean Square Error (RMSE): 0.67 | 1) Methodology is not explained well. 2) Dataset is small; data acquisition and splitting is not precise. 3) Cross-validation was applied. 4) Two models were compared.<br>5) other performance metrics were missing |
| [75] | To find the region of interest (ROI) of the fetal biometric and organs region in the US image | DCNN AlexNet | N/A | 16 to 27 weeks | ACC: 0.9043 | 1) Dataset acquisition and splitting is not apparent. 2) Validation is not applied. 3) Only accuracy as performance metrics |
| [78] | To detect of fetal abnormality in 2-D US images | ANN + Multilayered perceptron neural networks (MLPNN) | Gradient vector flow (GVF)<br>Median Filtering | 14 to 40 weeks | HC AC<br>MSE: 0.087 0.177 | 1) Dataset acquisition and splitting is not apparent. 2) Validation is not applied. 3) Only Mean square error as performance metrics |
| [79] | To develop a computer aided diagnosis and classification tool for extracting ultrasound sonographic features and classify IUGR fetuses | ANN | Two-Step Splitting Method (TSSM) for Reaction-Diffusion (RD) | various | RMSE: 0.91 to 0.94 | 1) Dataset acquisition and splitting is not apparent. 2) Validation is not applied. 3) Only RMSE as performance metrics |
| [76] | To develop an automatic classification algorithm on the US examination result using Convolutional Neural Network in Blighted | CNN | N/A | N/A | ACC: 0.60 | 1) Dataset acquisition is missing. 2) Insufficient dataset was used<br>3) Validation is not applied 4) Only accuracy as performance metrics |
| [77] | To proposes an intelligent system based on combination of ConvNet and PSO for Down Syndrome diagnosis. | CNN | Particle Swarm Optimization (PSO) | N/A | ACC: 0.99 | 1) Small dataset and dataset acquisition is not apparent. 2) Methodology could be explained in a better way 3) Only accuracy as performance metrics. 4) The idea is novel but obtained very high accuracy on small dataset is suspicious. |
| **Gestational age (GA)** | | | | | | |
| [80] | To automatic detect and measure the trans cerebellar diameter (TCD) in the fetal brain, which enables the estimation of fetal gestational age (GA) | CNN FCN | N/A | 16- to 26 weeks | ACC: 0.9789<br>IoU: 0.8162 | 1) Methodology and results are not clear. 2) Dataset is sufficient after augmentation, data acquisition and is precise. 3) No cross-validation was applied. 4) No comparison with state-of-the-art methods. 5) missing signification performance metrics |
| [81] | To accurately estimate the gestational age from the fetal lung region of US images. | U-NET | N/A | 24 to 40 weeks | Correlation coefficient: 0.76<br>P-value: 0.00001 | 1) Dataset acquisition is detailed but the dataset small, and no cross-validation is applied 2) Missing essential performance metrics. 3) Methodology was not clear for the reader. |
| [82] | To classify, segment, and measure several fetal structures for the purpose of GA estimation | U-NET<br>RESTNET | Residual UNET (RUNET) | 16th weeks | ACC: 0.93<br>IoU: 0.91<br>Mean Absolute Error: 1.89 centimeters | 1) Dataset acquisition and splitting is not apparent, and no cross-validation is applied 2) Methodology steps are clear and organized |
| **Gender identification** | | | | | | |
| [83] | To measure the accuracy of Learning Vector Quantization (LVQ) to classify the gender of the fetus in the US image" | ANN | Learning Vector Quantization (LVQ)<br>Moment invariants | N/A | Mean Absolute Error (MAE):0.3889.<br>Rec: 0.611<br>Pre: 0.58<br>F1: 0.556" | 1) Methodology is not clear, but obtained result is clearly explained. 2) Dataset is very small; data acquisition and splitting are not precise. 3) Cross-validation was applied, no augmentation. 4) Different models were compared. 5) Novel work by identifying gender in early stage of pregnancy. |

**Table 6.** Articles published using AI to improve fetus head monitoring: Objective, backbone methods, optimization, fetal age best result, and observations

| Study | Objective | Backbone Methods/Framework | Optimization/ Extractor methods | Fetal age | Best Obtained Result | observations |
|---|---|---|---|---|---|---|
| **Skull localization and measurement** | | | | | | |
| [87] | To localize the fetal head region in US imaging | Multi-scale mini-LinkNet network | N/A | 12 and 40 weeks | DSC: 0.9265<br>Difference (DF): 0.94 mm<br>Absolute Difference (ADF): 2.39 mm<br>HD: 3.53 mm | 1) methodology is clear. 2) result is blur for reader. 3) no cross validation is applied |
| [86] | To locate the fetal head from 3D ultrasound images using shape model | AdaBoost | Shape Model<br>Marginal Space<br>Haar-like features | 11 to 14 weeks | ACC: 0.7593 | 1) Methodology could be interpreted in better way. 2) Dataset very small. 3) Only accuracy as performance metrics. |
| [84] | To detect fetal head | Deep Belief Network (DBN)<br>Restricted Boltzmann Machines | Hough transform<br>Histogram Equalization | 11 to 14 weeks | ACC: 0.948<br>HD: 5.2033<br>Similarity Index (SI): 0.8472 | 1) Methodology is clear. 2) Only clinical dataset was reported and synthetic data were ignored. 3) Important performance metrics are missing. |
| [88] | To semantically segment fetal head from maternal and other fetal tissue | U-NET | Ellipse fitting | 12 and 20 weeks | ACC: 0.77<br>Average Error as DF: 14.96 mm | 1) methodology is clear. 2) Dataset very small even public dataset is available.<br>3) Important performance metrics are missing such as DSC |
| [103] | To automatically discover and localize anatomical landmarks. measure the HC, TV, and the TC | CNN | Saliency maps | 13 to 26 weeks | CSP LV/Cereb HC<br>TV alignment errors: 9.8 4.1<br>7.1 | 1) Methodology and results are confusing for the reader<br>2) Dataset acquisition and splitting are clear<br>3) Important performance metrics are missing |
| [95] | To demonstrate the effectiveness of hybrid method to segment fetal head | DU-Net | Scattering Coefficients (SC) | 13 to 26 weeks | DSC: 0.9733<br>HD: 1.58 mm | 1) Methodology could be explained better for the reader. 2) Dataset splitting is not apparent, and no validation. 3) Proposed method did not achieve the best result. |
| [96] | To segment fetal head using Network Binarization | Depthwise Separable Convolutional Neural Networks DSCNNs. | Network Binarization | 12 and 40 weeks | DSC: 0.9689 | 1)Methodology orientation is not structured. 2) no validation. 3) Important performance metrics are missing. |
| [97] | To segment the fetal skull boundary and fetal skull for fetal HC measurement | U-NET | Squeeze and Excitation (SE) blocks | 12 and 40 weeks | DSC: 0.9731<br>ADF: 2.69 mm | 1) Methodology orientation is clear. 2) Slightly improvement compared with original model. 3) Other performance metrics are missing |
| [98] | To automatic segment and estimate of HC ellipse. | Multi-Task network<br>based on Link-Net architecture (MTLN) | Ellipse Tuner | 12 and 40 weeks | DSC: 0.9684 ADF: 2.12 mm<br>DF: 1.13 mm HD: 1.72 mm | 1) Methodology and result are clear<br>2) Proposal method did not obtain the best result compared to the current work. |
| [99] | To capture more information with multiple-channel convolution from US images | Multiple-Channel and Atrous MA-Net | Encoder and Decoder Module | N/A | Dice: 0.973 Pre: 0.968 mm<br>Rec: 0.978 HD: 10.924 mm<br>ASD: 4.152 mm RMSD: 4.161 mm | 1) Methodology and result are clear and detailed. 2) dataset acquisition is clearly explained no validation is applied. 3) proposal method achieved the best result compared to other works except in precision. |
| [89] | To automatic segment fetal ultrasound image and HC biometry | Deeply Supervised Attention-Gated (DAG) V-Net | Attention-Gated Module | 12 and 40 weeks | DSC :0.9793 DF: 0.09 mm<br>ADF :1.77 mm HD: 1.27 mm | 1) Methodology and result are clear and detailed. 2) Dataset acquisition is clearly explained, no validation is applied 3) Proposal method achieved the best result compared to other state-of-the-art methods except in DSC. 4) Provided many comparisons with state-of-the-art methods |
| [90] | To compound a new US volume containing the whole brain anatomy | U-NET + Incidence Angle Maps (IAM) | CNN<br>Normalized Mutual Information (NMI) | 13 to 26 week | DSC: 0.811 AUC: 0.834 | 1) methodology and result are confusing for the reader. 2) dataset acquisition and splitting are not clearly defined. 3) paper organization is poor.4) no comparison with state-of-the-art methods. |
| [161] | To directly measure the head circumference, without having to resort the handcrafted features or manually labeled segmented images. | CNN regressor (Reg-Resnet50) | N/A | 12 to 40 weeks | MSE: 36.21 pixels MAE: 37.34 pixels<br>Huber Loss (HL): 38.18 pixels | 1) Methodology and result are simply explained and understandable. 2) Dataset acquisition and splitting are clearly defined. 3) Paper organization is poor. 4) First attempt to use CNN as a regressor to deal with fetal dataset without segmentation. |
| [105] | To propose region-CNN for head localization and centering, and a regression CNN for accurately delineate the HC | CNN regressor (U-net) | Tiny-YOLOv2 | 12 to 40 weeks | DSC: 0.9776 DF: 0.21 mm<br>ADF: 1.90 mm HD:1.32 mm | 1) Methodology and result are simply explained and understandable. 2. Dataset acquisition and splitting are clearly defined. 3) Second attempt to use CNN as a regressor to deal with fetal HC dataset without segmentation |
| [102] | To present a novel end to-end deep learning network to automatically measure the fetal HC, biparietal diameter (BPD), and occipitofrontal diameter (OFD) length from 2D US images | FCNN (SAPNet) | Regression network | 12 to 40 weeks | DSC: 0.9772 mIoU: 0.9646<br>ADF:2.03 mm HD: 0.47 mm<br>Mean Pixel ACC (mPA): 98.02 | 1) Methodology is clear, result has multiple comparisons confused the reader. 2) Dataset volume after augmentation were not reported. 3) regression network was not clarified |
| [104] | To segment fetal head from US images | FCN | Faster R-CNN | 12 to 40 weeks | DSC: 0.9773 DF: 1.49 mm<br>ADF: 2.33 mm HD: 1.39 mm | 1) Methodology is precise. 2) Dataset after augmentation is not reported. 3) No validation method, no comparison |
| [106] | To deals with a completely computerized detection device of next fetal head composition | Multi-Task network based on Link-Net architecture (MTLN) | Hadamard Transform (HT)<br>ANN Feed Forward (NFFE) Classifier | N/A | ACC: 0.97 | 1) Methodology and result is not precise. 2) Dataset acquisition and splitting are missing 3) No validation method, no comparison. 4) Only accuracy is reported 5) paper is not organized |
| [85] | To measure HC automatically | Random Forest Classifier | Haar-like features<br>ElliFit method | 18 to 33 weeks | DSC: 0.9666 RMSD: 1.77 mm<br>Pre: 0.9684 Rec: 0.9680<br>Spec: 0.9672 | 1) Methodology and result are precise. 2) Dataset is small with no augmentation and data acquisition is not detailed. 3) No validation method<br>4) Relay on prior knowledge of gestational age and scanning depth |
| [100] | To determine measurements of fetal HC and BPD | FNC | N/A | 18 to 22 weeks | HC Inter-expert Model<br>MAE 2.16 1.99<br>BPD<br>MAE 0.59 0.61<br>DSC 0.98 0.98 | 1) Methodology and result are precise<br>2) Dataset is large with no augmentation, and data acquisition is detailed<br>3) Comparison between expert annotation and model prediction |
| [163] | To segment the whole fetal head in US volumes | Hybrid attention scheme (HAS) | 3D U-NET + Encoder and Decoder architecture for dense labeling | 20 to 31 weeks | DSC: 0.9605 Jaccard Index (JI): 0.9242<br>HD boundaries: 4.60 mm | 1) Methodology and result are precise and detailed. 2) Dataset is small but augmented to obtain high volume. 3) Comparison is provided with another model. 4) Claimed to be the first investigation about whole fetal head segmentation in 3D US, but many works done before them, such as [101], [112]–[114] |
| [91] | To segment fetal head using a flexibly plug-and-play module called vector self-attention layer (VSAL) | CNN | Vector Self-Attention Layer (VSAL)<br>Context Aggregation Loss (CAL) | 12 and 40 weeks | DSC: 0.971 HD: 3.234<br>Pixel ACC (PA): 0.99 | 1) Methodology and result are precise and detailed. 2) Dataset is public; no augmentation is used. 3) Comparison is provided with another models. 4) Paper overwhelmed with unnecessary details |
| [101] | To provide automatic framework for skull segmentation in fetal 3D US | Two-Stage Cascade CNN (2S-CNN) U-NET | Incidence Angle Map<br>Shadow Casting Map | 20 to 36 weeks | DSC: 0.83 Jaccard index (JI): 0.70<br>Symmetric Surface Distance (SSD): 0.98 mm | 1) Methodology and result are precise. 2) Dataset is very small; the volume of data was not reported after the augmentation. 3) No validation method. 4) Claimed to be the first investigation about whole fetal head segmentation in 3D US. but many works done before them, such as [112]–[114] |
| [92] | To segment 2D ultrasound images of fetal skulls based on a V-Net architecture | Fully Convolutional Neural Network Combination (VNet-c) | N/A | 12 to 40 weeks | DSC:0.979 JI: 0.959<br>Pre: 0.976 ACC: 0.988 | 1) Methodology and result are precise. 2) The volume of data was not reported after the augmentation. 3) No validation method. 4) Comparative results were added. 5) Paper could be organized in better way |
| [93] | To segment the cranial pixels in an ultrasound image using a random forest classifier | Random Forest Classifier | Simple Linear Iterative Clustering (SLIC)<br>Haar Features | 25 to 34 weeks | Pre: 0.9901 ACC: 0.9722<br>Rec: 98.11 | 1) Methodology and result are not precise. 2) Dataset is very small with no augmentation, and data acquisition and splitting are not clear. 3) No comparison with other models 4) Paper is poorly written and inconsistent. |
| [94] | To automatically estimate fetal HC | U-Net | Monte-Carlo Dropout | 18 to 22 weeks | DSC: 0.982 HD: 1.29<br>ADF: 1.81 | 1) Methodology and result are not precise. 2) Dataset acquisition and splitting is not clear. 3) No validation method. |
| **Brain standard plane** | | | | | | |
| [107] | To automatically recognize six standard planes of fetal brains. | CNN+ Transfer learning DCNN | N/A | 18 to 22 weeks<br>40th week | ACC: 0.910 Pre: 0.855<br>Rec: 0.901 F1: 0.900 | 1) Clear methodology<br>2) Multi dataset are used with data augmentation |
| [114] | To help the clinician or sonographer obtaining these planes of interest by finding the fetal head alignment in 3D US | Random forest classifier | Shape model and template deformation algorithm<br>Hough transform | 19 to 24 weeks. | TC TV<br>Maximal error distance: 5.8<br>5.1 | 1) Methodology is not clear many objectives in one flow<br>2) Dataset very small<br>3) Only error rate as performance metrics |

**Table 6.** Continued

| Study | Objective | Backbone Methods/Framework | Optimization/ Extractor methods | Fetal age | Best Obtained Result | observations |
|---|---|---|---|---|---|---|
| [109] | To segment the fetal cerebellum from 2D US images. | U-NET +ResNet (ResU-NET-C) | N/A | 18 to 20 weeks | DSC: 0.87 HD: 28.15 mm<br>Pre: 0.90 Rec: 0.86 | 1) Methodology and results are presented clearly. 2) Dataset is small but acquisition, and splitting are clear. 3) Important performance metrics are used |
| [116] | To detect multiple planes simultaneously in challenging 3D US datasets | Multi-Agent Reinforcement Learning (MARL) | RNN<br>Neural Architecture Search (NAS)<br>Gradient-based Differentiable Architecture Sampler (GDAS) | 19 to 31 weeks | Dihedral angles between two planes (ANG): 9.75<br>difference between their ED towards the volume origin (DIS): 1.19<br>Structural Similarity Index (SSIM): | 1) Methodology and result detailed written<br>2) Second work that used of RL with the fetal dataset<br>3) Some details are not necessary to be included<br>4) Such as great work need future work<br>5) Some performance metrics are missing |
| [118] | To detect standard plane and quality assessment | Multi-task learning Framework Faster Regional CNN (MF R-CNN) | N/A | 14 to 28 weeks | Detection (mAP): 0.9404<br>Pre: 0.9776 Rec: 0.9785<br>Spec: 0.9833 F1: 0.9562<br>ACC: 0.9625 AUC:0.9889 | 1) Methodology and result are precise and detailed. 2) Dataset is transparent in terms of acquisition and splitting. 3) Classification module should adopt only one architecture instead of seven. 4) Detection only evaluated using mean average precision, where many metrics were used for classification |
| [119] | To tackle the automated problem of fetal biometry measurement with a high degree of accuracy and reliability | U-Net, CNN | Bounding-box regression (object-detection) | N/A | HC BPD<br>ACC: 0.9143 1<br>DSC 0.9539 | 1) Methodology and result are not precise<br>2) Dataset is very small but transparent in terms of acquisition and splitting<br>3) Only accuracy and DSC were used as performance evaluation |
| [117] | To determine the standard plane in US images | Faster R-CNN | Region Proposal Network (RPN) | 14 to 28 weeks | LS CP T CSP TV<br>AP 0.94 0.93 0.81 0.87 0.44 | 1) Methodology and result are precise. 2) Dataset acquisition is not detailed<br>3) No validation method. 4) Only Average Precision (AP) is reported |
| [112] | To address the problem of 3D fetal brain localization, structural segmentation, and alignment to a referential coordinate system | Multi-Task FCN | Slice-Wise Classification | 18 to 34 weeks | Alignment Error Average of HD: 9.3 mm<br>Brain Segmentation JI: 0.82<br>Classification Mean ACC: 0.964 | 1) Methodology and result are precise and detailed<br>2) Dataset is small with no augmentation<br>3) No comparison is provided for such a great work<br>4) Performance evaluation metrics should be used, such as AUC, DSC, F1 score |
| [113] | To simultaneously localize multiple brain structures in 3D fetal US | View-based Projection Networks (VP-Nets) | U-Net<br>CNN | 20 to 29 weeks | Average center deviation: 2.0 mm<br>Average scale deviation: 2.3 | 1) Methodology and result are precise and detailed. 2) Dataset is small, but augmentation volume was not reported. 3) Comparison is provided with other models |
| [108] | To automatically identify six fetal brain standard planes (FBSPs) from the non-standard planes. | Differential-CNN | Modified feature map | 16 to 34 weeks | ACC: 0.9311 Pre:0.9262<br>Rec: 0.9239 F1: 0.9239<br>AUC: 0.937 | 1) Methodology and result are precise. 2) After augmented the dataset splitting details is missing. 3) Comparative results were added<br>4) Paper was organized in a consistent way |
| [110] | To obtain the desired position of the gate and Middle Cerebral Artery (MCA) | MCANet | Dilated Residual Network (DRN)<br>Dense Upsampling Convolution (DUC) block | 28 to 40 weeks | DSC: 0.768 HD: 26.40 pixel<br>JI: 0.633 | 1) Some of related work were written in the methodology. 2) After augmented the dataset splitting details is missing. 3) Comparative results were added 4) First MCA dataset that should be public but not yet |
| [111] | To segment four important fetal brain structures in 3D US | Random Decision Forests (RDF) | Generalized Haar-features | 18 to 26 weeks | CP PVC CSP Cerebellum<br>DSC: 0.79 0.82 0.74 0.63 | 1) Methodology and result are not precise. 2) Dataset is very small with no augmentation, and data acquisition and splitting is not clear 3) No comparison<br>4) Paper is poorly written and inconsistent. |
| [115] | To automatically localize fetal brain standard planes in 3D US | Dueling Deep Q Networks (DDQN) | RNN-based Active Termination (AT) (LSTM) | 19 to 31 weeks | TC TT<br>Dihedral (ANG): 9.61 9.11<br>ED to origin (Dis): 3.40 2.66<br>SSIM: 0.69 0.70 | 1) Methodology and result are precise. 2) Dataset is small, acquisition and splitting is clear. 3) No validation method. 4) First work that use RL to localize fetal head |
| **Brain diseases** | | | | | | |
| [123] | To evaluate the feasibility of CNN-based DL algorithms predicting the fetal lateral ventricular width from prenatal US images. | ResNet50 | Faster R-CNN<br>Class Activation Mapping (CAM) | 22 to 26 weeks. | MAE of the predicted lateral ventricular :1.01 mm<br>ACC: 0.981 | 1) Methodology and result are confusing for the reader; no table or figure was seen to conclude the findings. 2) Dataset splitting is not well explained 3) Paper organization is poor. 4) No comparison with state-of-the-art methods |
| [120] | To recognize and separate the studied US data into two categories: healthy (HL) and hydrocephalus (HD) subjects. | CNN | N/A | 20 to 22 weeks. | ACC: 0.9720 Rec: 0.9795<br>Spec: 0.9823 | 1) methodology and result are explained and understandable. 2) dataset is minimal and has no augmentation 3) First attempt to identify hydrocephalus. |
| [124] | To automatic measurement of fetal lateral ventricles (LVs) in 2D US images. | Mask R-CNN | Feature Pyramid Networks (FPN)<br>Region Proposal Network (RPN) | N/A | ACC: 0.96 Pre: 0.9825<br>Rec:0.9491 Spec:0.9302 | 1) Methodology is clear. 2) Fetal GA was not reported. 3) No validation no augmentation. 4) First study proposing an automatic measurement method for [illegible] |
| [121] | To apply binary classification for central nervous system (CNS) malformations in standard fetal US brain images in axial planes. | CNN | Split-view Segmentation | 18 to 32 weeks | ACC: 0.963 Spec:0.959<br>Rec: 0.969 ROC:0.95 | 1) Methodology and result is clear. 2) Dataset is very big and transparent in terms of acquisition and splitting, but from the same center and many scanners were used.<br>3) No validation |
| [122] | To develop computer-aided diagnosis algorithms for five common fetal brain abnormalities. | Deep convolutional neural networks (DCNNs)<br>VGG-net | U-net<br>Gradient-Weighted Class Activation Mapping (Grad-CAM) | 18 to 32 weeks | Pre: 0.95 Rec: 0.96<br>F1: 0.95 | 1) Methodology and result are clear. 2) Dataset is very big and transparent in terms of acquisition and splitting, but from the same center and many scanners were used<br>3) is not the first attempt to develop algorithms that determine whether fetal brain images of standard axial neurosonographic planes (SANPs) are normal or abnormal.<br>4) The localization of lesions was currently based purely on the CAM. |

**Table 7.** Articles published using AI to improve fetus face monitoring: Objective, backbone methods, optimization, fetal age best result, and observations

| Study | Objective | Backbone Methods/Framework | Optimization/ Extractor methods | Fetal age | Best Obtained Result | observations |
|---|---|---|---|---|---|---|
| **Fetal facial standard plane (FFSP)** | | | | | | |
| [125] | To address the issue of recognition of standard planes (i.e., axial, coronal and sagittal planes) in the fetal ultrasound (US) image | SVM classifier | AdaBoost for detect region of interest, ROI)<br>Dense Scale Invariant Feature Transform (DSIFT)<br>Aggregating vectors for feature extraction fish vector (FV)<br>Gaussian Mixture Model (GMM) | 20 to 36 weeks | Mean Average Precision (mAP): 0.8995 | 1) Method and result are not clear. 2) Dataset is small, and the acquisition and splitting were not clear. 3) no data augmentation was not reported. 4) Other evaluation methods were not mentioned clearly |
| [126] | To automatically recognize the FFSP from US images | Deep convolutional networks (DCNN) | N/A | 20 to 36 weeks | ACC: 0.9699 Pre:0.9698<br>Rec: 0.9899 | 1) Method and result are clear. 2) Dataset is small, and acquisition was not clear. 3) No data augmentation was not reported. 4) No validation method |
| [127] | To automatically recognize FFSP via a deep convolutional neural network (DCNN) architecture | DCNN | t-Distributed Stochastic Neighbor Embedding (t-SNE) | 20 to 36 weeks | ACC: 0.9653 Pre:0.9698<br>Rec: 0.97 F1: 0.9699 | 1) Methodology and result is clear and detailed. 2) Dataset acquisition and splitting was clear. 3) Data augmentation and validation is employed 4) DCNN outperform |
| [128] | To automatic recognition of the fetal facial standard planes (FFSPs) | SVM classifier | Root scale invariant feature transform (RootSIFT)<br>Gaussian mixture model (GMM)<br>Fisher Vector (FV)<br>Principle Component Analysis (PCA) | 20 to 36 weeks | ACC: 0.9327 mAP: 0.9919 | 1) Multi-techniques were used in methodology and result is clear, detailed, and consistent. 2) Dataset acquisition and splitting was not clear. 3) Data augmentation and validation is not employed. 4) no comparison with other works. 5) For such great and detailed work, only precision and accuracy were reported. |
| [129] | To automatic recognize and classification of FFSPs | SVM classifier | Local Binary Pattern (LBP)<br>Histogram of Oriented Gradient (HOG) | 20 to 24 weeks | ACC: 09467 Pre:0.9427<br>Rec: 0.9388 F1:0.9408 | 1) Methodology and result are consistent. 2) Dataset splitting was not clear. 3) No data augmentation. 4) No comparison with other works. |
| **Face anatomical landmarks** | | | | | | |
| [130] | To detect position and orientation of facial region and landmarks | SFFD-Net (Samsung Fetal Face Detection Network) multi-class segmented | N/A | 14 to 30 weeks | L-Eye R-Eye Nose<br>L2 norm (mm): 2.21 1.66 1.31 | 1) Method and result are not clear. 2) Dataset is small, and the acquisition and splitting was not clear. 3) Volume after data augmentation was not reported 4) Other evaluation methods were not mentioned. |
| [131] | To detect landmarks in 3D fetal facial US volumes | CNN Backbone Network | Region Proposal Network (RPN)<br>Bounding-box regression | N/A | AP: 0.8250 mIoU:0.6377 | 1) Method and result are clear. 2) Dataset is small, and acquisition was not clear. 3) Volume after data augmentation was not reported. 4) No validation method. 5) Other evaluation methods were not mentioned " |
| [132] | To detect nasal bone for US of fetus | Back Propagation Neural Network (BPNN) | Discrete Cosine Transform (DCT)<br>Daubechies D4 Wavelet transform | 11 to 13 weeks | Nasal Bone Without<br>ACC:0.86 (43/50) 0.88(44/50) | 1) Methodology and result are inconsistent. 2. Dataset acquisition and splitting was not clear. 3) No data augmentation and validation are not employed 4) No comparison with other works. |
| **Facial expressions** | | | | | | |
| [133] | To recognize facial expressions from 3D US | ANN | Histogram equalization<br>Thresholding<br>Morphing<br>Sampling<br>Clustering<br>Local Binary Pattern (LBP)<br>Minimum Redundancy and Maximum Relevance (MRMR) | N/A | ACC: 0.91 Pre: 0.84<br>Rec: 0.85 | 1) In methodology, multi-techniques were used. However, they can be replaced by using CNN. 2) Dataset is small, and the acquisition was not clear. 3) Data augmentation was not reported. 4) No validation method |
| [134] | To recognize fetal facial expressions that are considered as being related to the brain development of fetuses | CNN | N/A | 19 to 38 weeks | ACC: 0.985 Spec: 0.993<br>Rec: 0.996 F1: 0.974 | 1) In methodology and result is clear. 2) Dataset acquisition was not clear. 3) volumes after data augmentation was not reported . 4) no validation method. 5) CNN outperform other method in [127] for a similar purpose. |

**Table 8.** Articles published using AI to improve fetus heart monitoring: Objective, backbone methods, optimization, fetal age best result, and observations

| Study | Objective | Backbone Methods/Framework | Optimization/ Extractor methods | Fetal age | Best Obtained Result | observations |
|---|---|---|---|---|---|---|
| **Heart disease** | | | | | | |
| [137] | To perform multi-disease segmentation and multi-class semantic segmentation of the five key components | U-NET + DeepLabV3+ | N/A | N/A | DSC:0.897 mIoU:0.720<br>Cross Entropy Loss: 0.034 | 1) Method and result are not clear. 2) dataset acquisition and splitting was not clear 3) no validation method 4) no comparison. |
| [139] | To recognize and judge Fetal congenital heart disease (FHD) development | DGACNN Framework | CNN<br>Wasserstein GAN + Gradient Penalty (WGAN-GP)<br>DANomaly<br>Faster-RCNN | 18–39 weeks | ACC: 0.850 AUC:0.881 | 1) Methodology and results are very detailed but confusing for the reader and could be written and organized consistent way. 2) Dataset is sufficient after the use of data augmentation. 3) Only ACC and AUC are reported. 4) Novel work and significant for clinical use. |
| [141] | To segment the ventricular septum in US | Cropping-Segmentation-Calibration (CSC) | YOLOv2 cropping module<br>U-NET segmentation Module<br>VGG-backbone Calibration Module | 18 to 28 weeks | mDSC:0.6950 mIoU: 0.5598 | 1) Methodology and results are clear. 2) Many techniques were used to build the framework. 3) Dataset acquisition splitting is clear. 4) No data augmentation. |
| [140] | To detect cardiac substructures and structural abnormalities in fetal US videos | Supervised Object detection with Normal data Only (SONO) | CNN<br>YOLOv2 | 18 to 34 weeks | Heart Vessels<br>mean AUC: 0.787 0.891<br>cardiac substructure detection<br>mAP: 0.702 | 1) Two techniques were integrated to build the framework. 2) Methodology and result are clear. 3) Large dataset and no augmentation needed. 4) Supplement materials were added, including real-time record. 5) Significant work and first application were seen in this survey. |
| [138] | To identify recommended cardiac views and distinguish between normal hearts and complex CHD and calculate standard fetal cardiothoracic measurements | Ensemble of Neural Networks | ResNet and U-Net<br>Grad-CAM | 18 to 24 week | ACC: 0.90 AUC:0.99<br>Rec: 0.95 Spec: 0.96<br>Negative predictive value in distinguishing normal from abnormal hearts: 100% | 1) Two modifies networks were used to build an ensemble framework 2) Methodology and results are very detailed, but a paper's organization is difficult for the reader. 3) Largest dataset in the field, acquisition, and splitting is clear 4) No comparison with other works. 5) Promising result was achieved. |
| [135] | To learn the features of Echogenic Intracardiac Focus (EIF) that can cause Down Syndrome (DS) whereas testing phase classifies the EIF into DS positive or DS negative based | Multi-scale Quantized Convolution Neural Network (MSQCNN) | Cross-Correlation Technique (CCT)<br>Enhanced Learning Vector Quantiser (ELVQ) | 24–26 weeks | ACC: 0.65 Rec: 0.70<br>Spec: 0.50 | 1) Methodology and result are not explained clearly. 2) Very small dataset, acquisition, and splitting was not clear. 3) No validation method, no data augmentation. 4) Evaluation metrics were not accurately reported. |
| [136] | To perform automated diagnosis of hypoplastic left heart syndrome (HLHS) | SonoNet (VGG16 ) | N/A | 18–22 weeks | AUC: 0.93 Pre: 0.83<br>Rec: 0.90 F1: 0.87 | 1) Methodology is based on other work [164] result is explained clearly 2) Large dataset but data acquisition, and splitting was not clear 3) No validation method, no data augmentation. 4) No comparison with other works. |

**Table 8**. Continued

| Study | Objective | Backbone Methods/Framework | Optimization/ Extractor methods | Fetal age | Best Obtained Result | observations |
|---|---|---|---|---|---|---|
| **Heart chamber's view** | | | | | | |
| [143] | To perform automated segmentation of cardiac structures | CU-NET | Structural Similarity Index Measure (SSIM) | N/A | DSC: 0.856 HD: 3.31<br>Pixel-level Accuracy (PA): 0.929 | 1) Methodology and results are clear. 2) No data augmentation and no validation method. 3) Dataset acquisition splitting is clear. |
| [144] | To accurate segment of seven important anatomical structures in the A4C view | DW-Net | Dilated Convolutional Chain (DCC) module<br>W-Net module based on the concept of stacked U-Net | n/a | DSC: 0.827 AUC:0.990<br>PA:0.933 | 1) Methodology and results are clear. 2) Two techniques were used to build the framework. 3) Dataset splitting is not clear. 4) No data augmentation<br>5) Claimed to be first time that a CNN based deep learning method is applied in the segmentation of multiple anatomical structures in A4C views. |
| [146] | To automatic quality control of fetal US cardiac four-chamber plane | Three CNN-based Framework | Basic-CNN, a variant of SqueezeNet<br>Deep-CNN with DenseNet-161 as basic architecture<br>The ARVBNet for real-time object detection. | 14 to 28 weeks | ACC Pre Rec F1<br>B-CNN: 0.9983 0.9982 0.9983 0.9983<br>D-CNN: 0.9951 0.9849 0.9877 0.9863<br>mAP of ARVBNet: 0.9352 | 1) Three different networks were used to build the framework. 2) Dataset splitting is not clear. 3) Each network was tested separately, and the average performance for the whole framework was not reported. |
| [147] | To localize the end-systolic (ES) and end-diastolic (ED) from ultrasound | Hybrid CNN based framework | YOLOv3<br>Maximum Difference Fusion (MDF)<br>Transferred CNN | 18 to 36 weeks | Pre Rec Spec F1<br>ES: 0.9569 0.9513 0.9427 0.9541<br>ED: 0.9490 0.9273 0.9547 0.9380<br>Average ACC: 0.9484 | 1) Many different techniques were integrated to build the framework. 2) Volume of the dataset after augmentation was not reported. 3) No validation method. 4) significant work and first-time MobileNet network were used to deal with the fetal US and achieved promising results. |
| [145] | To detect the fetal heart and classifying each individual frame as belonging to one of the standard viewing planes | FCN | N/A | 20 to 35 weeks | Segmentation error rate: 0.2348 | 1) Methodology and result are clear. 2) Large dataset and volume after augmentation is not reported. 3) Only error rate is reported and no validation method. |
| [142] | To jointly predict the visibility, view plane, location of the fetal heart in US videos. | Multi-Task CNN | Hierarchical Temporal Encoding (HTE) | 20 to 35 weeks | Classification Localization<br>ACC: 0.8358 0.7968 | 1) Methodology is not clearly explained. 2) dataset acquisition and splitting were not clear. 3) Only accuracy rate is reported, and no validation method |

**Table 9.** Articles published using AI to improve fetus abdomen monitoring: method details, objectives, health issue, dataset details and model observations

| Study | Objective | Backbone Methods/Framework | Optimization/ Extractor methods | Fetal age | Best Obtained Result | observations |
|---|---|---|---|---|---|---|
| **Abdominal anatomical landmarks** | | | | | | |
| [148] | To automatically detect two anatomical landmarks in an abdominal image plane stomach bubble (SB) and the umbilical vein (UV). | AdaBoost | Haar-like feature | 14 to 19 weeks | SB UV<br>ACC: 0.80 0.63<br>Spec: 0.70 0.65<br>Rec: 0.86 0.73<br>Error: 0.19 0.36 | 1) Methodology is explained clearly but result need more details. 2) Dataset acquisition and splitting was clear. 3) No validation method, no data augmentation. 4) No comparison with other works. |
| [149] | To localizes fetal abdominal standard plane (FASP) from US including SB, UV, and spine (SP) | Random Forests Classifier+ SVM | Haar-like feature<br>Radial Component-Based Model (RCM) | 18 to 40 weeks | Overall ACC: 0.816 | 1) Methodology and result are not clearly explained. 2) Dataset acquisition and splitting was not clear. 3) No validation method, no data augmentation. 4) Only detection rate was used to evaluate the performance. |
| [155] | To classify ultrasound images (SB, amniotic fluid (AF), and UV) and to obtain an initial estimate of the AC." | Initial Estimation CNN+U-Net | Hough transform | N/A | DSC AC measurement: 0.9255<br>Abdominal Plane acceptance ACC: 0.871 | 1) Many networks were used to build none integrated framework; methodology and results are confusing for the reader. 2) Dataset acquisition and splitting was not clear. 3) No validation method. 4) Evaluation metrics were not used clearly. 5) |
| [157] | To classify ultrasound images (SB, AF, and UV) and measure AC | CNN | Hough transform | 20 to 34 weeks | DSC AC measurement: 0.8919<br>Abdominal Plane acceptance ACC: 0.809 | 1) Many networks were used to build none integrated framework; methodology and results are confusing for the reader. 2) Dataset acquisition and splitting was not clear. 3) No validation method 4) Evaluation metrics were not used clearly. 5) |
| [150] | To find the region of interest (ROI) of the fetal abdominal region in the US image. | Fetal US Image Quality Assessment (FUIQA) | L-CNN is able to localize the fetal abdominal ROI AlexNet<br>C-CNN then further analyzes the identified ROI<br>DCNN to duplicate the US images for the RGB channels | 16 to 40 weeks | SB UV<br>AUC: 0.999 0.996<br>ACC: 0.990 0.983 | 1) Many networks were used to build the FUIQA framework.<br>2) Methodology and results are confusing for the reader.<br>3) Dataset splitting was not clear<br>4) No validation method<br>5) No comparison |
| [151] | To localizes the fetal abdominal standard plane from ultrasound | Random forest classifier+ SVM classifier | Radial Component-based Model (RCM)<br>Vessel Probability Map (VPM)<br>Haar-like features | 18 to 40 weeks | SB UV SP ROI<br>AUC: 0.99 0.99 0.99 0.98<br>Overall ACC: 0.856 | 1) Two classifiers were used to improve detection speed and accuracy. 2) Methodology is not consistently explained. 3) Dataset splitting was not clear after data Augmentation. 4) No validation method |
| [156] | To diagnose the (prenatal) US-images by design and implement a novel framework | Defending Against Child Death (DACD) | CNN<br>U-Net<br>Hough-man transformation | N/A | Seg class<br>SB UV SB UV<br>ACC: 0.991 0996<br>Overall ACC: 99.7 | 1) Classification integrated with segmentation was used to improve detection speed and accuracy. 2) Methodology is not consistently explained. 3) Dataset splitting was not precise, no augmentation. 4) No validation method. 5) Some inconsistency during the comparison with other works, however, DACD provide promising result. |
| [152] | To detect important landmarks employed in manual scoring of ultrasound<br>Images. | AdaBoost | Haar-like feature | 18 to 37 weeks | SB UV<br>ACC: 0.7894 0.6280<br>AUC: 0.80 0.57 | 1) Methodology and result is clear but not detailed 2) Dataset acquisition and splitting are missing; no augmentation. 3) No validation method. 4) No comparison with other works. |
| [154] | To automatically select the standard plane from the fetal US volume for the application of fetal biometry measurement. | AdaBoost | One Combined Trained Classifier (1CTC)<br>Two Separately Trained Classifiers (2STC)<br>Haar-like feature | 20 to 28 weeks | Rec: 0.9129 Pre: 0.7629 | 1) Methodology is clear but result mossing more detailed 2) Dataset acquisition and splitting are missing. 3) No validation method. 4) No comparison with other work. 5) First work in the area. |
| [153] | To localize the FASP from US images. | DCNN | Fine Tuning with Knowledge Transfer<br>Barnes-Hut Stochastic Neighbor Embedding (BH-SNE) | 18 to 40 weeks) | ACC: 0.910 | 1) Methodology is not clear. 2) Dataset acquisition and splitting are precise, and data augmentation was reported. 3) No validation method. 4) Comparison with other works 5) Only accuracy was reported. |

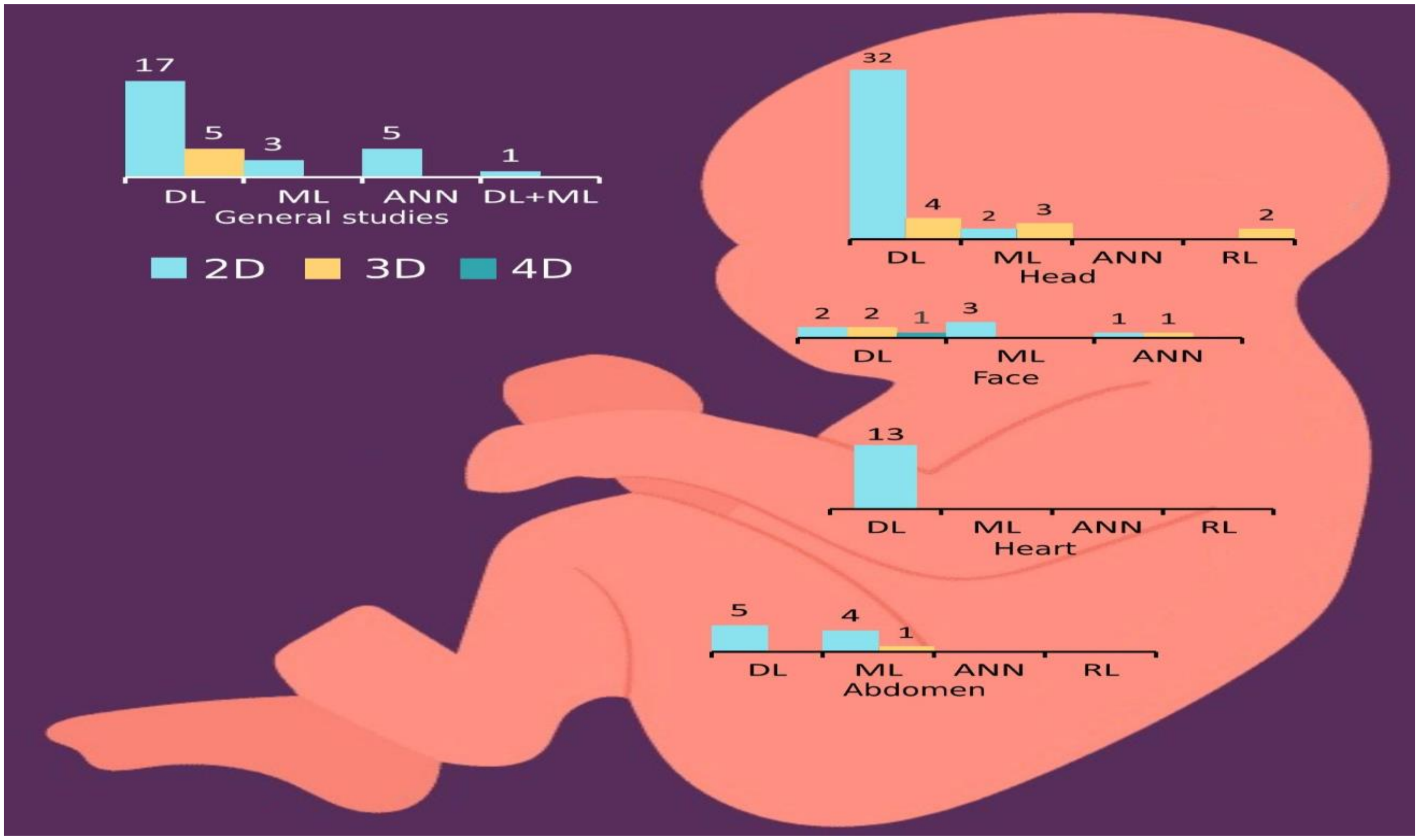

**Figure 4**. AI techniques that used within all studies in this survey. AI techniques are ordered by US images type and further categorized by fetal organ. Totals here equal the number of included papers.

# 6. Discussion

## 6.1 Principal findings

AI techniques applied are shown in Figure 4. Deep learning (DL) was the most utilized techniques, seen in (n=81, 75.70%) studies. Traditional Machine leaning (ML) models are used in (n=16, 14.95%) studies. Artificial neural network (ANN) models are only used in (n=7, 6.5%) studies. We also found the use of Reinforcement learning (RL) in (n=2, 1.86%) studies. Both deep learning as well as machine learning models were used in one study.

- **Deep learning for fetus health**

As seen in Figure 4. DL was used heavily in all fetal organ experiments including general fetus issues, fetus head, fetus face, fetus heart, and fetus abdomen. Furthermore, as seen in most of the models (n=74, 69.1%), Convolutional Neural Network (CNN) used as a backbone of deep learning and is used in all tasks including classification, segmentation and object detection [54], [56], [57], [59], [63], [65]–[68], [72], [73], [76], [77], [81], [82], [84], [87]–[91], [93]–[99], [102]–[107], [109], [110], [113], [117]–[124], [126], [127], [130], [131], [134]–[144], [146], [147], [150], [153], [155]–[157], [161], [163]. Additionally, Fully Connected Neural Networks (FCNNs) is similarly largely utilized as a backbone or part of the framework (n=11, 10.28%) of DL, as seen in [63], [65], [70], [71], [80], [92], [100], [104], [112], [145], [146]. In addition to CNN and FCN, a Recurrent Neural Network (RNN) used to perform classification tasks in hybrid frameworks as seen in (n=5, 4.67%) studies [54], [57], [59], [60], [70]. U-Net is a convolutional neural network that was developed for biomedical image segmentation. Therefore, we found U-NET to be the most frequently utilized segmentation model as a baseline of the framework or as optimization. This was seen in (n=23, 21.50%) [63], [73], [81], [82], [88], [90], [94], [95], [97], [101], [109], [113], [118], [119], [122], [137], [138], [141], [143], [144], [155], [156], [163]. Other segmentation models were also used, such as DeepLabV3 in [137], LinkNet in [106], and the Encoder-Decoder network based on VGG16 in [72]. A Residual Neural Network (ResNet) was used to perform some tasks and enhance the framework efficiency as seen in (n=6, 5.60%) studies [62], [67], [82], [109], [110], [123], [138]. Some popular object detection models (n=12, 11.21%) were utilized to perform detection tasks in the fetus head, face, and heart. Additionally, Region Proposal Network (RPN) with bounding-box regression technique was used in [119], [131]. Faster R-CNN was utilized in [104], [117]–[119]. Further, YOLO (You Only Look Once) models were utilized; YOLOv2 in [105], [140], [141] and YOLOv3 in [147]. Mask R-CNN was utilized only in [124] and Aggregated Residual Visual Block (ARVB) was used only in [146]. Surprisedly, we have found CNN is used for regression to predict the accurate measurement of HC as seen in [102], [103], [105], [161]. We found that, due to the noise and characteristics of US images, DL still relies on traditional feature extraction techniques in image analysis used to optimize the DL frameworks for better performance. For example, Hough transform was utilized with head and abdomen US images as seen in [84], [155]–[157]. Additionally, image processing techniques such as histogram equalization was used in [84]. Besides these tools, Class Activation Maps (CAM) are a helpful classification tool in computer vision that was utilized for brain visualization in [122], [123], fetal part structure in [54], and heart chamber view in [138]. DL shows promising results for the identification of certain fetus diseases as seen in the identification of heart diseases [135]–[141], brain diseases [120]–[124], and growth diseases [75]–[77]. Table 10 list of DL studies exhibiting technical novelty that achieved promising results in each category.

**Table 10.** Best DL study in each category based on the achieved result

| Mian Organ | Sub-section | Best study |
|---|---|---|
| Fetal Body | Fetal part structures | [62], [65] |
| | Anatomical structures | [67], [73] |
| | Growth disease | [77] |
| | Gestational age | [81] |
| Head | Skull localization | [105], [161] |
| | Brain Standard plan | [107], [108], [110] |
| | Brain disease | [120], [124] |
| Face | Fetal facial standard | [127] |
| | Face anatomical | [130] |
| | Facial expression | [134] |
| Heart | Heart disease | [135], [137], [138], [140] |
| | Hear chamber view | [144], [147] |
| Abdomen | Abdominal anatomical | [155], [156] |

- **Machine leaning for fetus health**

Most of the research utilizing machine learning for fetal health monitoring was conducted between 2010 and 2015, as seen in (n=14, 13.8%). Only two studies [85], [129] were published in 2018 and 2021, respectively. We found that ML models were used for all fetus organs except for the heart and identification of fetus disease. The most used ML algorithm was Random Forest (RF), as seen in (n=8, 7.47%) studies. RF was used as a baseline classifier in [85], [93], [111], [114]. RF was used with other classifiers as well, with Support Vector Machine (SVM) in [149], [151] and Probabilistic Boosting Tree (PBT) in [55]. Simple Linear Iterative Clustering (SLIC) was also used in [64]. The SVM classifier was used in (n=6, 5.60%) studies. SVM was used as a baseline classifier in [53], [125], [128], [129] and utilized with other classifiers as seen in [149], [151]. AdaBoost was used in (n=5, 4.67%) studies as a baseline classifier in [86], [148], [152], [154] and to detect the region of interest (ROI) in [125]. Moreover, ML traditional algorithm still rely on Haar-like features (digital image features used in object recognition) while being utilized with US images as seen in (n=9, 8.41%) studies [85], [86], [93], [111], [148], [149], [151], [152], [154]. Gaussian Mixture Model (GMM) and Fisher Vector (FV) were used in (n=3, 2.80%) studies, where GMM is used to simulate the distribution of extracted characteristics throughout the images. Then, FV represents the gradients of the features under the GMM, with respect to the GMM parameters as seen in [53], [118], [125].

DL and ML models were seen in [58]. the SVM with decision fusion for classification and features extracted by fine-tuning AlexNet. Finally, we found the first research to utilize ML model in monitoring fetus health was in [154], which used to classify abdominal anatomical landmarks.

- **Artificial Neural Network for fetus health**

ANN is the cut edge between ML and DL [165]. As seen in Figure 4, ANN was used in (n=7, 6.5%) studies [69], [74], [78], [79], [83], [132], [133]. ANN models were used for the identification of growth diseases [74], [78], [79], fetus gender [83], facial expressions [133], face anatomical landmarks (Nasal Bone) [132], and anatomical structures (nuchal translucency (NT)) [69]. Hence, ANN was not utilized to identify brain or heart structures, nor accompanying disease. Finally, we concluded that the first utilization of ANN to monitor fetus health was in 2010 [69]; the main goal was to detect NT, which helps early identify Down syndrome. The second utilization of ANN was completed in 2011 [132] to detect nasal bone. This detection also helps for early identification of Down syndrome.

- **Reinforcement learning for fetus health**

The first attempt to utilize RL to monitor fetus health was in 2019 [115]. Dueling Deep Q Networks (DDQN) was employed with RNN-based active termination (AT) to identify brain standard planes (trans-thalamic (TT) and trans-cerebellar (TC)). The second time RL was utilized to monitor fetus health was reported on in a recent study from 2021 [116]. A multi-agent RL (MARL) was used in a framework that compromises RNN and includes both neural architecture search (NAS) as well as gradient-based differentiable architecture sampler (GDAS). This framework achieved promising results for identifying the brain standard plane and localize mid-sagittal, transverse (T), coronal (C) planes in volumes, trans-thalamic (TT), trans-ventricular (TV), and trans-cerebellar (TC).

### 6.1.1 Strengths

To the best of our knowledge, this survey is the first to explore all AI techniques implemented to provide fetus health care and monitoring during different pregnancy trimesters. This survey may be considered comprehensive as it does not focus on specific AI branches or diseases. Rather, it provides a holistic view of AI's role in monitoring fetuses via ultrasound images. This survey will benefit both readers, first medical professionals to have sight about the current AI technology in a development stage and assist data scientists in overcoming existing research and implication problems. secondly, for data scientists to continue future investigating in developing lightweight models and overcoming problems that slow the transformation of this technology based on a solid guideline proposed by the medical professionals. It may be considered a robust and high-quality survey. Because the most prominent health and information technology databases were searched using a well-developed search query, the search was sensitive and exact. Lastly, the utilization of techniques like scanning grey literature databases (Google Scholar), biomedical literature database (PubMed, Embase) World's leading citation databases (Web of Science), Subject Specific Databases (PsycINFO), and world's leading source for scientific, technical, and medical research (ScienceDirect, IEEE explore, ACM Library) indicates that this study's risk of publication bias is low.

### 6.1.2 Limitations

A comprehensive systematic survey was conducted. However, the search within the databases was conducted between 22nd and 23rd June 202, so we might miss some new studies. Besides that, the words "pregnancy" or "pregnant" or "uterus" were not used as search terms to identify relevant papers. Therefore, some fetus health monitoring studies might have been missed, lowering the overall number of studies. Additionally, this survey focused more on a clinical rather than the technical; therefore, some technical details may have been missed. Lastly, only English studies were included in the search. As a result, research written in other languages was excluded.

## 6.2 Practical and research implication

The diagnosis of US imaging plays a significant role in clinical patient care. Deep learning, especially the CNN-based model, has lately gained much interest because of its excellent image recognition performance. If CNN were to live up to its potential in radiology, it is expected to aid sonographers and radiologists in achieving diagnostic perfection and improving patient care [166]. Diagnostic software based on artificial intelligence is vital in academia and research and has significant media relevance. These systems are largely based on analyzing diagnostic images, such as X-RAY, CT, MRI, electrocardiograms, and electroencephalograms. In contrast, US imaging, which is non-invasive, non-expensive, and non-ionizing, has limited AI applications compared to other radiology imaging technologies [167].

This survey found one available AI-based application to identify fetal cardiac substructures and indicate structural abnormalities as seen in [140]. From these findings, many studies achieved promising results in diagnosing fetus diseases or identifying specific fetus landmarks. However, to the best of our knowledge there is no randomized controlled trial (RCT) or pilot study carried out at a medical center or any adaptation of an AI-based application at any hospitals. This hesitance could be due to the following challenges [167]; 1) The present AI system can complete tasks that radiologists are capable of but may make mistakes that a radiologist would not make. For example, radiologists may not see undetectable modifications to the input data. These changes may not be seen by human eyes but would nevertheless impact the result of AI system's categorization. In other words, a tiny variation might cause a deep learning system to reach a different result or judgment. 2) Developers require a specific quantity of reliable and standardized data with an authorized reference standard to train AI systems. Annotated images may become an issue if this is done through retroactive research. Datasets may also be difficult to get as the firms that control them want to keep them private and preserve their intellectual property. Validation of an AI system in the clinic can be difficult as it frequently necessitates multi-institutional collaboration and efficient communication between AI engineers and radiologists. Validating an AI system is also expensive and time-consuming. Finally, 3) when extensive patient databases are involved, ethical and legal concerns may arise.

This survey found some of these challenges were addressed in some studies. For example, applying transfer learning on AI model learned natively on US images and fine-tuning the model on an innovative data set collected from a different medical center and/or different US equipment. Another way to address the challenges of a small data set is to employ data augmentation (e.g., tissue deformation, translations, horizontal flips, adding noise, and images enhancement) to boost the generalization capacity of DL models. It is recommended that data augmentation settings should be carefully chosen to replicate ultrasound images' changes properly [168]. Advanced AI applications are already being used in breast and chest imaging. Large quantities of medical images with strong reference standards are accessible in various fields, allowing the AI system to be trained. Other subspecialties like fetus disease, musculoskeletal disease or interventional radiology are less familiar with utilizing AI. However, it seems that in the future, AI may influence every medical application that utilizes any images [169].

This survey found that no standard guideline is being followed or developed that focused on AI in diagnostic imaging to meet clinical setting, evaluation, and requirements. Therefore, the Radiology Editorial Board [170] has created a list of nine important factors to assist in evaluating AI research as a first step. These recommendations are intended to enhance the validity and usefulness of AI research in diagnostic imaging. These issues are outlined for authors, although manuscript reviewers and readers may find them useful as well: 1) Carefully specify the AI experiment's three image sets (training, validation, and test sets of images). 2) For final statistical reporting, use a separate test set. Overfitting is a problem with AI models, which means they only function for the images they are trained to recognize. It is ideal to utilize an outside collection of images (e.g., the external test set) from another center. 3) Use multivendor images for each step of the AI assessment, if possible (training, validation, test sets). Radiologists understand medical images from one vendor are not identical to theirs; radiomics and AI systems can help in identifying changes in the images. Moreover, multivendor AI algorithms are considerably more interesting than vendor-specific AI algorithms. 4) Always justify the training, validation, and test sets' sizes. Depending on the application, the number of images needed to train an AI system varies. After just a few hundred images, an AI model may be able to learn image segmentation. 5) Train the AI algorithm using a generally recognized industry standard of reference. For instance, chest radiographs are interpreted by a team of experienced radiologists. 6) Describe any image processing for the AI algorithm. Did the authors manually pick important images, or crop images to a limited field of view? How images are prepared and annotated has a big impact on radiologist's comprehension of the AI model. 7) Compare AI performance against that of radiology professionals. Competitions and leader boards for the "best" AI are popular among computer scientists working in the AI field. The area under the receiver operating characteristic curve is often used to compare one AI to another (AUC). On the other hand, physicians are considerably more interested in comparing the AI system to expert readers, when treating a patient. Experienced radiologist readers are recommended to benchmark an algorithm intended to identify radiologic anomalies. 8) Demonstrate the AI algorithm's decision-making process. As previously mentioned, computer scientists working on imaging studies often describe their findings as a single AUC/ACC number. This AUC/ACC is compared to the previous best algorithm, which is a competitor. Unfortunately, the AUC/ACC value has little bearing on clinical medicine on its own. Even with a high AUC/ACC of 0.95, there may be an operating mode in which 99 out of 100 anomalies are overlooked. Many research teams overlay colored probability maps from the AI on the source images to assist doctors in comprehending the AI performance. 9) The AI algorithm should be open source so performance claims may be independently validated; this is already a prevalent recommendation, as this survey found only (n=9, 8.4%) studies [62], [70], [92], [105], [138]–[141], [163] made their works publicly available.

## 7. Future work and conclusion

Throughout this survey, various ways that AI techniques have been used to improve fetal care during pregnancy have been discussed, including: 1) determining the existence of a live embryo/fetal and estimate the pregnancy's age, 2) identifying congenital fetus defects and determining the fetus's location, 3) determining the placenta's location, 4) checking for cervix opening or shortening and evaluating fetal development by measuring the quantity of amniotic fluid surrounding the baby, 5) evaluating the fetus's health and see whether it is growing properly. Unfortunately, due to both ethical and privacy concerns as well as the reliability of AI decisions, all aspects of these models are not yet utilized as end applications in medical centers.

Undergraduate sonologist education is becoming more essential in medical student careers; consequently, new training techniques are necessary. As seen in other medical fields, AI-based applications and tools were employed for medical and health informatics students [171]. Moreover, AI-mobile-based applications were useful for medical students, clinicians, and allied health workers alike [172]. To the best of our knowledge, there are no AI-based mobile applications available to assist radiologist or sonographer students in medical college. Besides that, medical students may face many challenges during fetal US scanning [173]. Therefore, our future work will propose an AI- mobile-based application to help radiology or sonography students identify features in fetus US image and answer the following research questions: 1) How is the lightweight model efficient and accurate when localizing the fetal head, abdomen, femur, and thorax from the first to the third trimester?; 2) How efficient and accurate the lightweight model will identify the fetal gestational age (GA) from the first to the third trimester?; 3) How will inter-rater reliability tests validate the proposed model and compare the obtained result with experts and students using the intra-class correlation coefficient (ICC) [174]?.

We have concluded that AI techniques utilized US images to monitor fetal health from different aspects through various GA. Out of 107 studies, DL was the most widely used model, followed by ML, ANN, and RL. These models were used to implement various tasks, including classification, segmentation, object detection, and regression. We found that even the most recent studies rely on the 2D US followed by 3D, and that 4D is rarely used. Furthermore, we found that most of the work was targeted the fetus head followed by the body, heart, face, and abdomen. We identified the lead institutes in this field and their research.

This survey discusses the availability of the dataset for each category and highlighted their promising result. In addition, we analyzed each study independently and provided observations for the future reader. All optimization and feature extraction methods were reported for each study to highlight the unique contribution. Moreover, DL novel and unique works were highlighted for each category. The research and practical implications were discussed, and prospective research and clinical practitioner recommendations were provided. Finally, a future research direction has been proposed to answer the gap in this survey.

**Key Points**

- Artificial intelligence and machine learning methods can be successfully employed to optimize fetus care.
- Deep learning methods are the most common approach to monitor fetal health ,and the fetal head analysis is gaining traction.
- Future research should focus on employing these technologies and developing an actual application that can be used at medical colleges and centers.

## Supplementary data

Multimedia Appendix 1: Search strings

Multimedia Appendix 2: Data extraction form

## Data Availability Statement

Available upon request

## Funding

No funding

## Conflicts of Interest

None declared